\documentclass[a4paper,fleqn]{cas-sc}

\usepackage[authoryear,longnamesfirst]{natbib}
\usepackage{graphicx}
\usepackage{amsmath}
\usepackage{amssymb}
\usepackage{booktabs}
\usepackage{multirow}
\usepackage{subcaption}
\usepackage{threeparttable}
\usepackage{placeins}
\usepackage{caption}
\usepackage{soul}
\title[mode = title]{Towards Explainable Conversational AI for Early Diagnosis with Large Language Models}
\shortauthors{Tabassum et al}
\shorttitle{}

\author[1]{Maliha Tabassum}[type=author,orcid=0009-0000-2502-3637]
\ead{maliha2154901035@student.bup.edu.bd}

\author[2]{M. Shamim Kaiser}[type=author,orcid=0000-0002-4604-5461]
\ead{mskaiser@juniv.edu}

\affiliation[1]{organization={Department of Information and Communication Technology},
    addressline={Bangladesh University of Professionals},
    city={Dhaka},
    postcode={1216},
    country={Bangladesh}}
\affiliation[2]{organization={Institute of Information Technology},
    addressline={Jahangirnagar University},
    city={Savar},
    postcode={1342},
    country={Bangladesh}}

\begin{document}
\maketitle
\begin{abstract}
Healthcare systems around the world are grappling with issues such as inefficient diagnostics, rising costs, and limited access to specialists. These challenges often contribute to delays in treatment and poorer health outcomes. Most existing AI and deep learning–based health assessment systems offer limited interactivity and transparency, reducing their usefulness for user-centered health support. This research introduces a conversational chatbot powered by a Large Language Model (LLM), using GPT-4o, Retrieval-Augmented Generation, and explainable AI techniques. The chatbot engages users in a dynamic conversation to extract and normalize symptoms while identifying and ranking potential health conditions through similarity matching and adaptive questioning. Using Chain-of-Thought prompting, the system also provides more transparent explanations of its reasoning process. When evaluated against traditional machine learning models, including Naive Bayes, Logistic Regression, SVM, Random Forest, and KNN using both TF-IDF and CountVectorizer feature extraction, the proposed LLM-based system achieved a Top-1 accuracy of 90\% and a Top-3 accuracy of 100\%. The system was additionally evaluated through a cross-sectional expert evaluation involving 17 physicians across all 14 conditions, with the results indicating generally favorable assessments of conversational quality, early diagnostic plausibility, and safety-related criteria. These findings demonstrate the potential of explainable conversational AI as a health and well-being support tool for early symptom assessment. However, the proposed system is not intended for clinical diagnosis or clinical decision-making, and further validation would be required before any use in healthcare practice.

\end{abstract}

\section{Introduction}
Health and well-being remain among the most significant global challenges. Key issues include limited access to timely health information, high costs of healthcare services, and shortages of healthcare professionals in many countries. These problems can delay appropriate care-seeking and contribute to preventable deterioration in well-being. In Bangladesh, the situation is similarly concerning. As an example, in 2022, people spent roughly 61 USD per capita on healthcare\footnote{\url{https://data.worldbank.org/indicator/SH.XPD.CHEX.PC.CD?locations=BD}}, yet health spending was only 2.39 percent of GDP\footnote{\url{https://data.worldbank.org/indicator/SH.XPD.CHEX.GD.ZS?locations=BD}}. Almost 44 USD per capita of this was paid directly out of pocket by individuals\footnote{\url{https://data.worldbank.org/indicator/SH.XPD.OOPC.PC.CD?locations=BD}}. This implies that the majority of families are forced to pay for care with their own resources, placing a heavy financial burden on many individuals.

Both communicable and non-communicable diseases impose a substantial global burden. In 2021, the number of cases and deaths caused by malaria was estimated to be 263 million cases and 597000 deaths globally with 95 percent of the fatalities reported in Africa \cite{WHO2024Malaria}. Mycobacterium tuberculosis causes Tuberculosis (TB), which in 2023 had afflicted almost 10.8 million people and caused nearly 1.25 million deaths \cite{WHO2024TB}. By April 2024, dengue fever, an infectious condition spread by mosquitoes, had reached a record high of more than 7.6 million cases (3.4 million confirmed), and more than 3 000 deaths, already surpassing the 2023 figures \cite{WHO2024Dengue}. Chronic hepatitis B infection is still prevalent, and it is estimated that there are approximately 254 million individuals with the virus in 2022 and that over 1.1 million deaths are linked to the infection each year elsewhere \cite{WHO2023HBV}. Asthma was one of the respiratory diseases with an incidence of approximately 262 million individuals and a death burden of approximately 455000 in 2019 \cite{WHO2019Asthma}. Digestive diseases are also common: gastro-esophageal reflux disease (GERD) had an estimated global prevalence of approximately 825.6 million cases in 2021 \cite{GBD2021GERD} and irritable bowel syndrome (IBS) has a pooled world-wide prevalence of about 14.1 \% \cite{Ford2020IBS}. The most prevalent nutritional disorder is iron-deficiency anemia, which is experienced by approximately 1.92 billion individuals (approximately 24.3 percent of the global population) in 2021 \cite{GBD2021Anaemia}. Acute diarrheal Illness resulted in approximately 1.7 billion childhood episodes and 444000 deaths in children under five years of age per year \cite{WHO2020Diarrhoea}. Mental disorders are equally heavy burdens: depression impacts almost 280 million individuals \cite{WHO2019Depression} and anxiety disorders impact nearly 301 million \cite{WHO2019Anxiety}. Seasonal viral fevers like influenza cause a burden of about 1 billion cases per year, 3-5 million severe cases, and 290-650 thousand deaths, with over 775 million confirmed cases and 7 million deaths as of mid-2024 being due to COVID-19 \cite{WHO2024COVID} compared to 775 million cases and 7 million deaths per year. Lastly, neurological diseases including migraine were present in approximately 1.16 billion individuals in 2021, up 58 percent since 1990 \cite{GBD2021Migraine}. All of these numbers highlight the ongoing and pervasive nature of global health issues.

The lack of affordable and accessible healthcare demonstrates the necessity of intelligent solutions. Despite recent advancements, significant gaps remain in symptom assessment methodologies. High-performance deep learning models remain largely uninterpretable. Although they achieve strong predictive accuracy, they function as black-box systems and provide limited insight into the rationale behind their decisions. Their application in conversational health-support settings has not been sufficiently explored, even with the integration of explainable AI techniques. There is limited focus on conversational health-support contexts. Most existing research emphasizes structured data processing rather than the dynamic, iterative nature of user--system interactions. While recent studies based on large language models have begun to address this limitation, further expansion is required. Dynamic knowledge integration remains inadequate. Many systems rely on static datasets and lack mechanisms for continuous updating with new medical knowledge, thereby limiting their adaptability to evolving health conditions and symptom variations. Current symptom-assessment systems also lack adaptive questioning mechanisms that respond to contextual condition information. Instead, they typically employ fixed question sets that do not adjust dynamically to the evolving assessment context.

Motivated by these gaps, this study proposes an explainable conversational health-support system built on large language models and retrieval‑augmented generation. Unlike traditional machine learning approaches that rely on fixed symptom matching and opaque predictions, the proposed method enables free‑text symptom interpretation, adaptive questioning, and user‑visible reasoning throughout the assessment process. In comparison to prior LLM‑based health-support systems, our approach integrates real‑time confidence updates, explicit symptom attribution, and test‑based refinement, enhancing transparency and user trust. While the system demonstrates strong assessment performance, it also inherits limitations associated with LLM dependency and synthetic evaluation settings, which we discuss in later sections.

The primary objectives of this research are as follows: (a) to design a conversational, user-friendly LLM-based diagnostic chatbot capable of supporting early symptom assessment through the accurate identification and ranking of potential health conditions; (b) to incorporate explainable AI mechanisms that enhance transparency and interpretability by providing clear reasoning for the chatbot's responses and condition assessments; (c) to evaluate the performance of the proposed LLM-based system in conversational symptom-assessment scenarios by comparing it with traditional machine learning models; and (d) to assess the system's relevance, explainability, usability, and potential utility through a structured expert evaluation conducted by licensed physicians.

This research begins with a review of existing studies on artificial intelligence-based health-support methods and the datasets used in healthcare. Next, the background technologies, tools, methodology, and system architecture of the proposed approach are presented. The results are then discussed, with a focus on system performance, explainability, and comparisons with traditional machine learning approaches. Finally, the study concludes with the key findings and outlines potential directions for future research.

\section{Related Work}

\subsection{Recent ML/DL Approaches}

In this review, we focus primarily on studies that perform text- or structured-data-based disease diagnosis. Traditional machine learning (ML) and deep learning (DL) approaches have been widely applied for disease prediction and classification, using structured datasets or limited textual representations. 

Rahman et al. \cite{Rahman2025} applied feature selection with LightGBM and XGBoost to predict diabetes from the Pima Indian Diabetes Dataset and the more recent DiaHealth dataset, achieving high predictive accuracy while reducing input features for interpretability. In a breast cancer study, \cite{Tachicart2025} compared SVM, logistic regression, random forest, and ANN on a small, imbalanced dataset, highlighting the superior performance of ensemble and neural network models in classifying malignant and benign cases. Nazir et al. \cite{Nazir2023} focused on heart disease prediction using structured laboratory data from the Framingham Heart Study dataset, implementing decision trees, SVM, random forest, and ANN to identify patients at risk with reasonable accuracy. For textual data, a multiclass EHR classification study \cite{Zhang2026} applied TF–IDF representations with logistic regression, SVM, Multinomial Naive Bayes, and random forest to 9,633 clinical notes across four disease categories, demonstrating the potential of classical ML methods in clinical NLP. Finally, an ensemble DL + ML model \cite{Park2021} combined a two-layer neural network with LightGBM and XGBoost to predict 39 emergency-room diseases from 5,145 cases and over 326,000 laboratory tests, using SHAP values to interpret feature importance and achieve robust disease classification. Table \ref{tab:MLDLStudies} summarises the key ML/DL studies discussed above, including their methodologies and reported performance metrics.

While these studies demonstrate strong predictive performance within their specific domains, they primarily rely on structured inputs or predefined text representations, have limited ability to handle varied free-text symptom descriptions, and do not provide interactive natural-language explanations.

\begin{table}[htbp]
\centering
\caption{Summary of Key ML/DL Assessment Studies}
\label{tab:MLDLStudies}
\begin{tabular}{p{3cm} p{4cm} p{4cm} p{3.5cm}}
\toprule
\textbf{Reference} & \textbf{Domain} & \textbf{Methodology} & \textbf{Performance} \\
\midrule
\cite{Rahman2025} & Diabetes Prediction & Feature selection (RFE, GWO, PSO, GA, Boruta) + LightGBM & LightGBM achieved 85.16\% accuracy, 85.41\% F1-Score \\
\cite{Tachicart2025} & Breast Cancer ML Comparison Study & SVM, Logistic Regression, Random Forest, ANN & RF: F1 = 88.41\% (best); ANN: F1 = 86.96\% \\
\cite{Nazir2023} & Heart Disease Prediction & Decision Tree, Random Forest, SVM, ANN & SVM achieved 86.53\% accuracy (best) \\
\cite{Zhang2026} & Multiclass EHR Text Classification & TF–IDF + Logistic Regression, SVM, MNB, Random Forest & Logistic Regression: Accuracy = 0.8469, Macro F1 = 0.8339 (best) \\
\cite{Park2021} & 39 Disease Prediction & Two-layer DNN + LightGBM + XGBoost ensemble using 86 lab features & 92\% accuracy, 81\% F1-Score (top 5 diseases) \\
\bottomrule
\end{tabular}
\end{table}

\subsection{LLM-Based Approaches}

Tu et al. \cite{tu2024conversationaldiagnosticai} proposed AMIE, a multi-turn health-support dialogue conversational assessment large language model (LLM). Their system outperformed primary care physicians in symptom-assessment accuracy across all k values ($P < 0.05$), while being rated superior on 30 of 32 specialist-evaluated health-support axes. Nevertheless, AMIE was not interpretable at a fine-grained level, which restricted the understanding of how single symptoms affected the assessment results, which is essential in clinician trust and decision-making. Med-PaLM \cite{singhal2022largelanguagemodelsencode} is a medical adaptation of Google PaLM based on health-related datasets including MedQA and PubMedQA and fine-tuned on them. Their work made significant progress in the field of medical question answering, delivering 67.6\% accuracy. However, AMIE provides limited fine-grained interpretability, making it difficult to trace how individual symptoms contribute to specific assessment outcomes, which is important for user trust and decision-making. Rekha et al. \cite{rekha2024} compared a small language model (distilgpt2) and a larger language model (gpt2) to evaluate their predictive accuracy and resource demands when mapping disease symptoms to potential diagnoses.

Gupta et al. \cite{gupta2025digitaldiagnostics} compared GPT-4 and other LLMs in symptom-based condition assessment with promising accuracy. Although this research confirmed the assessment opportunities of foundation models, it was based on zero-shot prompting, which limits its use in real-world health-support settings. MDAgents \cite{mdagents2024} is a system in which a set of LLMs work together to complete a complex assessment problem. This method proved to be promising in the process of coordinating various models of differential diagnosis. Nonetheless, it operates within a static MCQ framework, lacking real-world conversational, iterative, and user-centric health-support interaction.  Toma et al. \cite{clinicalcamel2023} applied a dialogue-based LLM to encode clinical knowledge to predict diseases. Whereas the model had high benchmark performance and outperformed GPT-3.5, it was still prone to hallucinations. Table \ref{tab:llm_methods} summarises the LLM-based approaches discussed above, highlighting their methodologies and reported performance.

Collectively, these studies demonstrate the rapid development of LLMs in healthcare. However, challenges related to explainability, conversational adaptability, and real-time symptom-guided assessment interaction remain insufficiently explored.

\begin{table}[htbp]
\centering
\caption{LLM-Based Assessment Methods}
\label{tab:llm_methods}
\begin{tabular}{p{3cm} p{4cm} p{4cm} p{3cm}}
\toprule
\textbf{Reference} & \textbf{Domain} & \textbf{Methodology} & \textbf{Performance} \\
\midrule
\cite{tu2024conversationaldiagnosticai} & Cardiovascular, Respiratory, and Internal Medicine & Self-Play, PaLM 2, fine-tuning  &Higher top-k diagnostic accuracy than PCPs (p < 0.05)\ \\
\cite{singhal2022largelanguagemodelsencode} & Medical Question Answering & PaLM (540B parameters), Flan-PaLM & 67.6\% accuracy on MedQA \\
\cite{gupta2025digitaldiagnostics} & Medical Diagnosis & GPT-4, 4o, 3.5, Gemini, o3 & GPT-4	precision=0.96 and recall=0.92 \\
\cite{mdagents2024} & Medical Diagnosis & Multi-agent LLM collaboration& Achieved SOTA results in 7/10 benchmarks \\
\cite{clinicalcamel2023} & Clinical case studies & Dialogue-Based Knowledge Encoding & 53.2\% (Step 1) and 58.2\% (Step 3) \\
\bottomrule
\end{tabular}
\end{table}

\section{System Architecture and Methodology}

\subsection{Background Tools \& Technologies}

The health-support platform is built using advanced libraries and frameworks that are oriented to process medical information, language translation, and machine learning. These technologies ensure system stability, efficiency, and accuracy in the symptom assessment process.

\subsubsection{RAG Architecture}

\begin{figure}[pos=htbp]
\includegraphics[width=0.9\textwidth]{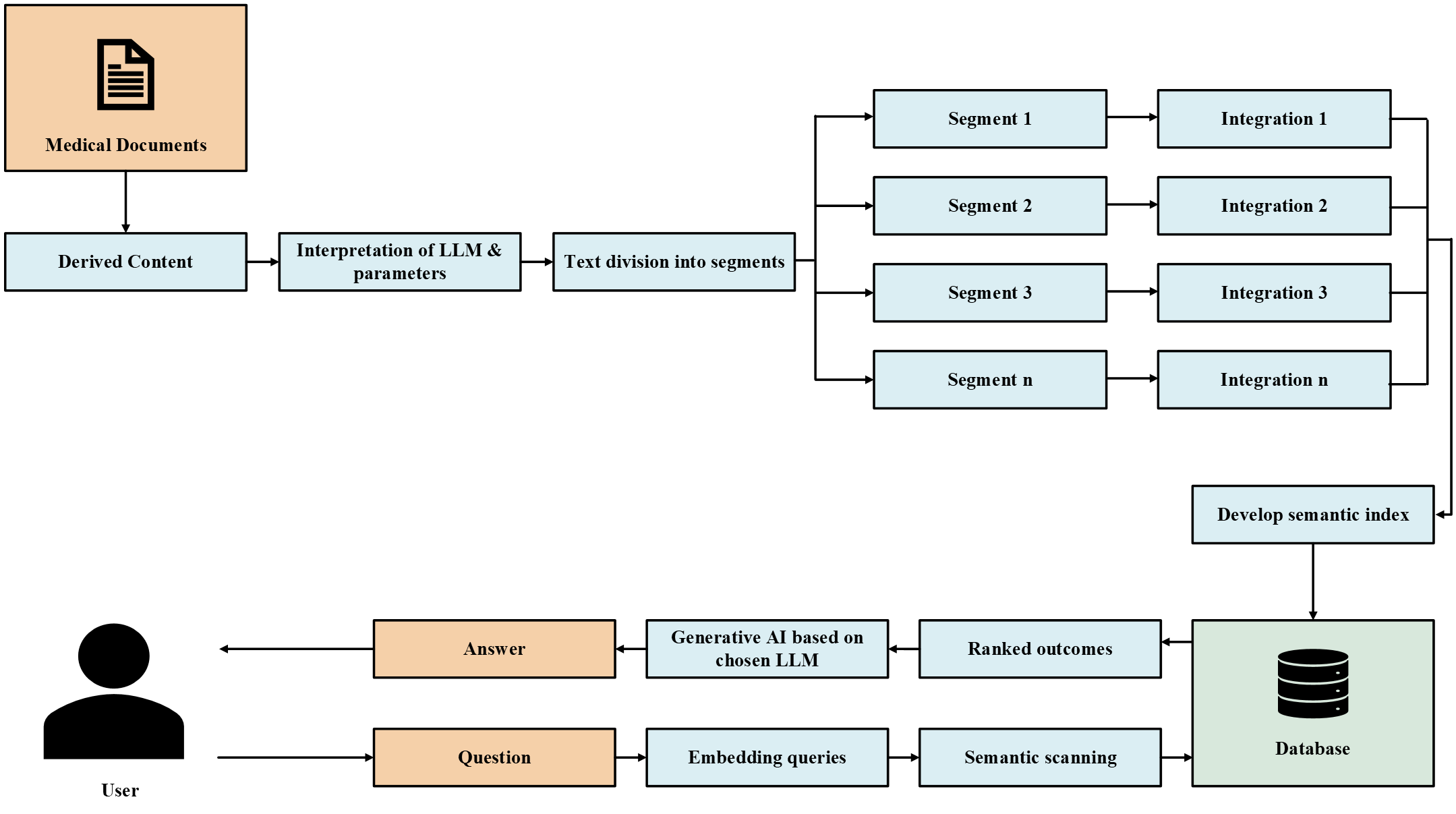}
\centering
\caption{General workflow of a Retrieval-Augmented Generation (RAG) system}
\label{rag}
\end{figure}

Our system is based on a Retrieval-Augmented Generation (RAG) architecture, which combines the strengths of retrieval-based systems and generative AI models. Figure \ref{rag} illustrates the general RAG workflow, which serves as the foundation of the proposed system. In a typical RAG pipeline, documents are chunked and converted into embeddings stored in a vector database for semantic retrieval. At query time, the user input is embedded, similar chunks are retrieved via similarity search, and the top results are used as context for an LLM to generate a grounded response.

Our proposed system implements this architecture for early symptom assessment and health guidance. FAISS is used as the vector database to efficiently retrieve relevant medical knowledge, while Azure OpenAI models generate context-aware responses. The RAG framework integrates the retrieved health information with user-reported symptoms and applies AI-driven reasoning to produce preliminary health-support outputs based on the retrieved context.

\subsubsection{Azure OpenAI Services and LangChain Framework}

\textbf{AzureChatOpenAI}: This serves as the main interface, providing access to OpenAI's GPT-4o model through Azure's cloud infrastructure. The system ensures consistency with a temperature setting of 0.3 and keeps the output concise with a token limit of 50. There are built-in retry mechanisms to handle potential API failures, ensuring system reliability in health-support applications \cite{azure_openai_docs}.

\textbf{AzureOpenAIEmbeddings}: This component converts medical text into numerical vectors, which allows matching user inquiries with the applicable medical knowledge with precision. It helps the system understand the context of user symptoms and match them with medical jargon\cite{azure_embeddings_docs}.

\textbf{LangChain Framework}:
\begin{itemize}
    \item \textbf{LLMChain}: The core orchestration module, combining user input with medical knowledge to generate ranked health-support outputs and maintain consistent assessment protocols \cite{langchain_docs}.
    \item \textbf{PromptTemplate}: Helps structure the language model's output by ensuring that responses are limited to condition names, avoiding unnecessary explanations and maintaining clarity in the assessment process \cite{langchain_prompts_docs}.
    \item \textbf{RecursiveCharacterTextSplitter}: This tool segments lengthy medical documents into manageable chunks, maintaining context and allowing effective information retrieval \cite{langchain_splitters_docs}.
\end{itemize}

\subsubsection{Document and Data Processing}

\textbf{FAISS (Facebook AI Similarity Search)}: This vector storage and retrieval tool allows for efficient matching of user queries with relevant medical knowledge, ensuring fast and accurate responses from the system, even with large datasets \cite{faiss_docs}.

\textbf{PyPDF2}: This library aids in extracting text from PDFs with complex layouts, enhancing the document processing pipeline and supporting various medical document formats \cite{pypdf2_docs}.

\textbf{Scikit-learn Metrics}: We used Scikit-learn's metrics module to evaluate system performance. It helps measure accuracy, precision, recall, and F1-score, providing valuable insights for continuous improvement of the assessment model \cite{sklearn_metrics_docs}.

\textbf{Pandas and NumPy}: Pandas manages the data pipeline, tracking user data, assessment results, and metrics \cite{pandas_docs}. NumPy facilitates efficient numerical computations, especially for operations involving similarity searches and embeddings\cite{numpy_docs}.

\textbf{Matplotlib}: This library was used to visualize the system's performance, allowing us to compare the effectiveness of LLM-based health-support methods with traditional machine learning models \cite{matplotlib_docs}.

\subsection{System Overview}

\begin{figure}[pos=htbp]
\centering
\includegraphics[width=0.9\textwidth]{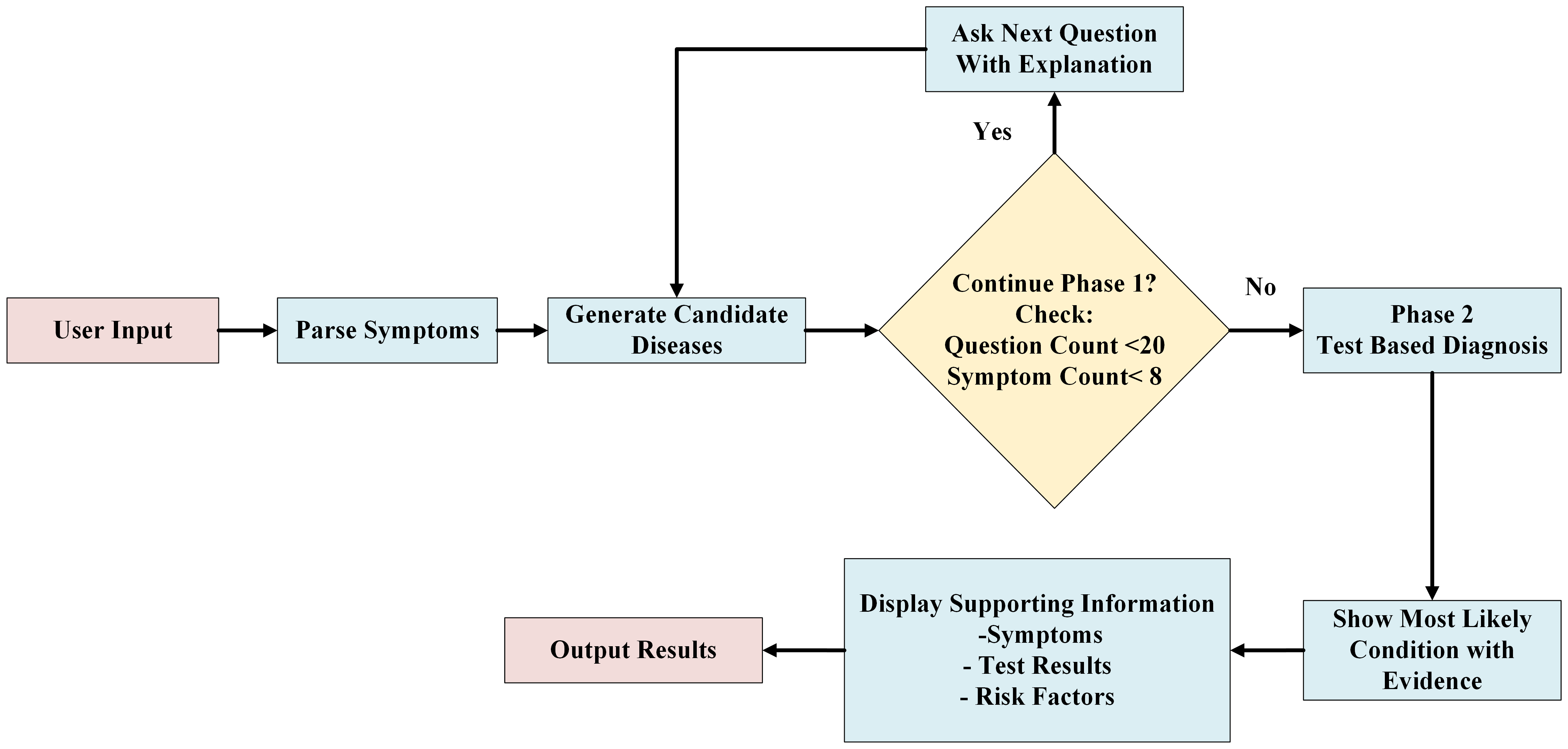}
\caption{End-to-End Workflow of the Proposed Health-Support System}
\label{fig:architecture}
\end{figure}

Our health-support system consists of two stages. The first stage is symptom-based assessment, in which the system requests the user to provide information about their symptoms and interprets the data to create a ranked set of likely conditions. Once sufficient information has been gathered, the second step proceeds to the next stage, where the system requests relevant laboratory test results concerning the related conditions and eventually provides the most probable conditions according to the user response. Figure \ref{fig:architecture}     illustrates the workflow of the proposed health-support system, including symptom elicitation, test-based refinement, evidence presentation, and final assessment generation. Each component of the system has been described. 

\begin{enumerate}
    \item \textbf{User Input and Symptom Parsing:}
The system accepts free-text symptom descriptions from the user. The input is tokenized and passed through a symptom extraction module that performs keyword normalization and synonym mapping. Extracted symptoms are converted into a structured representation for downstream retrieval and scoring.
    \item \textbf{Candidate Condition Generation:} The structured symptom set is embedded into a dense vector representation using a pre-trained embedding model. A similarity search is then performed against a disease--symptom knowledge base stored in a vector database. The system retrieves and ranks candidate conditions based on cosine similarity scores between the user symptom vector and condition vectors.
    \item \textbf{Phase 1: Symptom Elicitation:} The system performs iterative symptom clarification using a ranked condition list. It identifies missing or low-confidence symptom attributes and generates follow-up questions to refine the symptom profile. If the number of confirmed symptoms is below a defined threshold (e.g., fewer than 8), the system continues elicitation to improve assessment confidence and reduce ambiguity in the embedding space. 
    \item \textbf{Phase 2: Test-Based Diagnosis:} Once the system reaches sufficient assessment confidence, it selects the top three candidate conditions and retrieves their associated recommended laboratory tests from the knowledge base. The system then queries the user for corresponding test results and incorporates this additional structured information into the final scoring function to refine condition ranking. 
    \item \textbf{Display Supporting Information:} The system aggregates contributing evidence for each candidate condition, including matched symptoms, confirmed or rejected features, and relevant test results. This information is used to compute and present a justification summary for the ranked predictions. 
    \item \textbf{Final Assessment with Evidence:} The system outputs the most probable condition along with the top-ranked alternatives. Each prediction is accompanied by supporting evidence derived from symptom similarity scores and test result alignment, enabling interpretability of the final assessment.
\end{enumerate}

\subsection{Knowledge Base Construction}

A comprehensive medical knowledge base was constructed covering 14 conditions across four categories. These diseases include Malaria, Tuberculosis, Dengue Fever, Hepatitis B, Asthma, Gastroesophageal Reflux Disease (GERD), Irritable Bowel Syndrome (IBS), Iron-Deficiency Anemia, Acute Diarrheal Illness, Depression, Anxiety Disorders, Viral Fever, Covid-19 and Migraine. The diseases included in the knowledge base were selected based on their global prevalence, public health significance, and representation of diverse communicable and non-communicable health conditions, as discussed in Section 1.

The knowledge base was constructed by scraping information from authoritative publicly accessible sources, including guidelines and reports published by the World Health Organization (WHO) and other established medical organizations \cite{WHO_Malaria_FactSheet,WHO_Malaria_Topic,WHO_Malaria_QA,WHO_Malaria_Guidelines,WHO_TB,WHO2024Dengue,CDC_HepatitisB,WHO2023HBV,WHO_HepatitisB_Guidelines,WHO2019Asthma,NCBI_GERD,WGO_GERD,RomeIV_IBS,NCBI_IBS,WHO_Anemia,WHO2020Diarrhoea,WHO2019Depression,WHO_Depression_Topic,WHO2019Anxiety,CDC_ViralFever,Mayo_ViralIllness,WHO_COVID19,ICHD3_Migraine,ICHD3_Pocket}.

\begin{figure}[pos=htbp]
\centering
\includegraphics[width=0.9\textwidth]{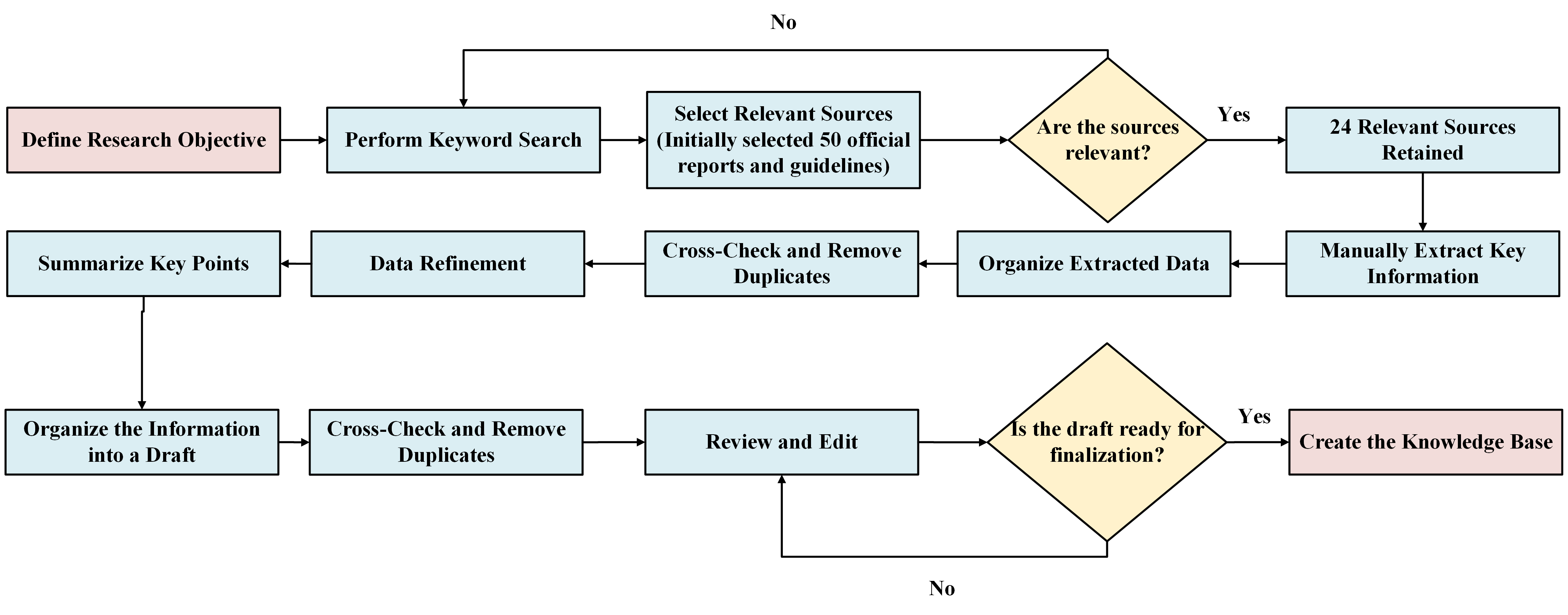}
\caption{Knowledge Base Construction Pipeline}
\label{data-collection}
\end{figure}

Figure \ref{data-collection} illustrates the pipeline used for knowledge base construction, including data acquisition, extraction, and structuring of medical knowledge.

For each disease, between 6 and 10 clinically relevant symptoms were collected from these sources. Symptoms were manually reviewed, standardized, and linked to common linguistic variations through synonym mapping to improve symptom recognition during user interaction. In addition to symptoms, associated laboratory tests, risk factors, and supporting assessment information were included when available. The resulting knowledge base provides structured condition profiles that support symptom retrieval, condition ranking, and test-based assessment refinement.

Table \ref{tab:diseases} shows the distribution of diseases in our knowledge base.

\begin{table}[htbp]
\centering
\caption{Condition Categories in Medical Knowledge Base}
\label{tab:diseases}
\begin{tabular}{|l|c|}
\hline
\textbf{Condition Category} & \textbf{Number of Conditions} \\
\hline
Infectious Conditions & 6\\
\hline
Chronic Conditions & 4 \\
\hline
Mental Health & 2 \\
\hline
Other Common Conditions & 2 \\
\hline
Total & 14 \\
\hline
\end{tabular}
\end{table}

One of the most effective methods for measuring similarity between sets is the Jaccard similarity coefficient. Since each condition is represented by a set of associated symptoms, Jaccard similarity provides a useful measure for quantifying symptom overlap across diseases.

\begin{figure}[pos=htbp]
\centering
\includegraphics[width=0.9\textwidth]{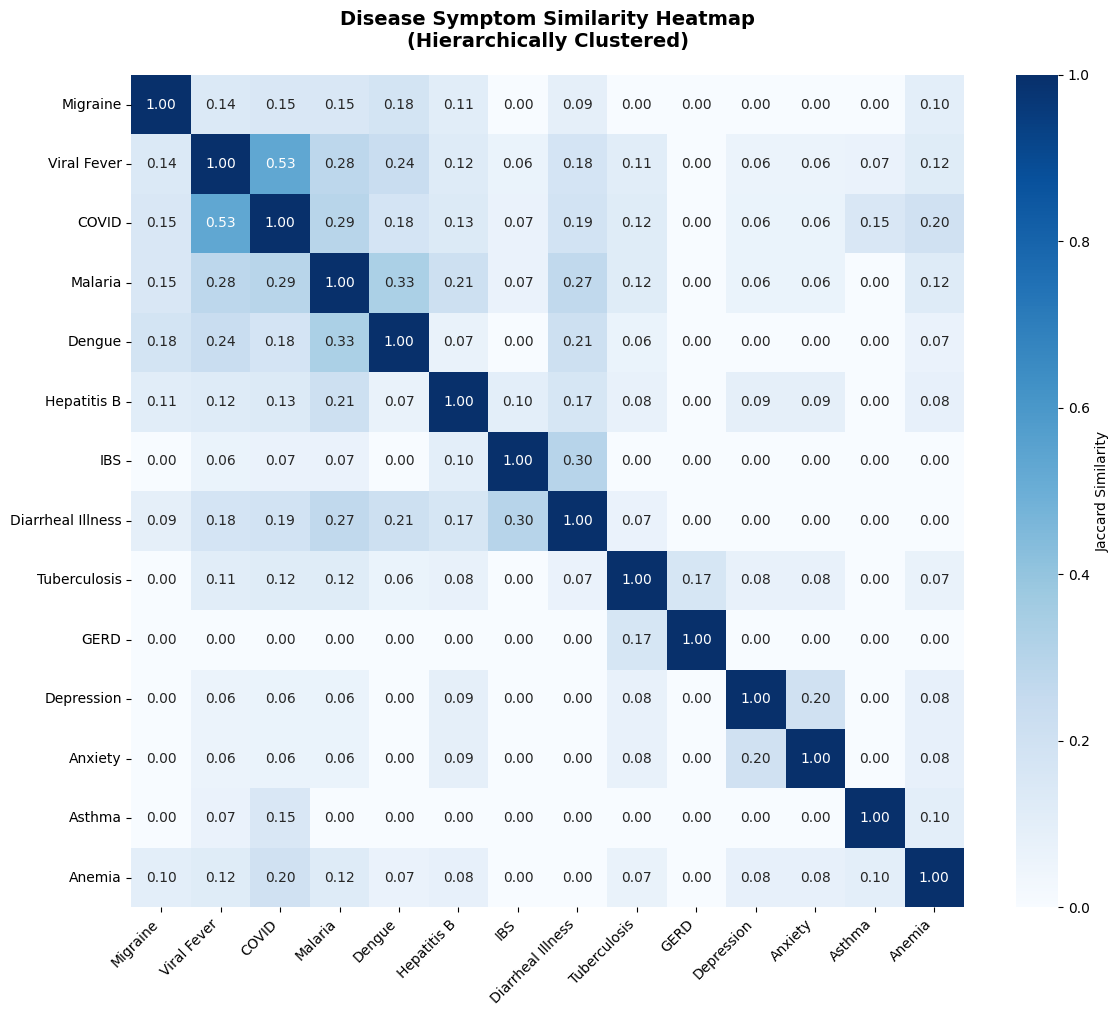}
\caption{Heatmap of pairwise symptom similarity between conditions using Jaccard similarity coefficients. Rows and columns represent the 14 conditions included in the knowledge base, while each cell indicates the proportion of shared symptoms between a condition pair. Higher values correspond to greater symptom overlap.}
\label{fig:symptom_heatmap}
\end{figure}

Symptom similarity analysis using Jaccard similarity coefficients revealed significant overlap between certain conditions. Figure \ref{fig:symptom_heatmap} presents pairwise symptom overlap across diseases using Jaccard similarity. The highest overlap occurs between Viral Fever and COVID-19 (0.53), reflecting their shared systemic symptoms. All other pairs show substantially lower similarity, suggesting the knowledge base maintains sufficient inter-condition discriminability to support reliable diagnosis.

The knowledge base contains four types of information for each disease:
\begin{enumerate}
\item \textbf{Symptoms:} Standardized descriptions of clinical manifestations, weighted by severity and assessment significance.
\item \textbf{Risk Factors:} Demographic, environmental, and lifestyle factors associated with increased condition risk.
\item \textbf{Threshold Values:} Quantitative assessment criteria from laboratory tests and clinical measurements.
\item \textbf{Management Information:} Evidence-based treatment recommendations and referral guidelines.
\end{enumerate}

While the current knowledge base supports effective assessment reasoning within the selected condition scope, it remains limited in size and diversity. As a result, the system may exhibit reduced robustness when evaluated on unseen cases that differ significantly from the constructed dataset, particularly in terms of symptom variation, comorbid conditions, and natural language expression. However, the retrieval-augmented architecture allows the system to scale to larger and more diverse datasets without requiring modifications to the underlying inference mechanism, making it adaptable to broader real-world distributions.

\subsubsection{Data Preprocessing}
After collecting the data from PDFs and online sources, the data undergo a cleaning and organization process to prepare them for use by the chatbot.

The content is split into primary sections which are symptoms, risk factors, and assessment criteria, with the help of regular expressions. This renders the data of each and every condition well organized. Afterward, the symptoms are given different levels of weight according to severity. The greater the severity of a symptom, such as chest pain or fatigue, the greater its impact on the assessment and the higher the priority assigned by the chatbot to more critical conditions.

After a thorough cleaning and structuring of the data, it is saved in a Pandas DataFrame. Such arrangement allows searches and retrieval of pertinent information with ease particularly when combined with RAG system that assists the chatbot in locating the appropriate information in dialogue.

\subsubsection{Integration into the System}
The RAG system is linked with the knowledge base, which enables the chatbot to mention significant data in the course of chatting with users. The system integrates a pre-trained language model with a knowledge retrieval system, supporting the accuracy of the assessment process.

In order to conduct semantic searches, the disease-related information is converted into vectors using embedding models and stored in FAISS. This serves the purpose of the chatbot to make it easy to find the most recent information in correspondence to the queries of the user. The information is also transformed into the Sentence-Transformer embedding representing semantic meanings of the symptoms, the risk factors, and assessment criteria. This enhances the skill of the chatbot in aligning the question of the users with the knowledge base.

\subsection{Preprocessing of Input Data}
The preprocessing of the user input is an important stage of data analysis preparation. This step converts raw data into a structured format that will be utilized by the chatbot. It is then tokenized into smaller units (such as words or phrases) to make it easier to analyze the text. For instance, the sentence "I have a headache and a fever" becomes \texttt{["I", "have", "a", "headache", "and", "a", "fever"]}.

Next, common words like "the", "and", and "is" (called stopwords) are removed because they do not contribute significant meaning. This reduces unnecessary data and makes the analysis more efficient. Finally, normalization is done by converting all text to lowercase and removing irrelevant characters, like punctuation, ensuring everything is consistent and easier to work with.

\subsection{Document Processing and Retrieval System}

The RAG system handles medical documentation through several steps. PyPDFLoader is used to extract text from medical PDFs. The documents are then divided into blocks of 1000 characters with 200-character overlap in order to facilitate processing. Afterward, the system creates the text in the form of vectors with the help of sentence-transformers. These embeddings are compiled in FAISS, and can be quickly accessed when interacting with the user.

The retrieval system is based on converting user queries into embeddings with the same model as document chunks. The top-k most relevant document chunks are retrieved. The context retrieved assists the LLM to produce the right response.

\subsection{Conversation Management System}

\subsubsection{Symptom Tracker}

The symptom tracker serves as the memory of the consultation, which retains both confirmed and rejected symptoms reported by the user as well as the record of questions that have been posed. Denied symptoms decrease the probability of diseases, computed using the formula:
\begin{equation}
    \tau_s = -0.8 \times w_s
\end{equation}
where \( w_s \) is the original assessment weight of the symptom. The condition score is updated dynamically as:
\begin{equation}
    \text{Score}(d) = \sum_{\text{matched } s} \alpha_s \cdot w_s - \sum_{\text{denied } s} \tau_s
\end{equation}
where \( \alpha_s = 1.0 \) for exact matches and \( \alpha_s = 0.6 \) for semantic matches.

\subsubsection{Dynamic Question Generation}

The selection of questions is guided by the ability to distinguish between diseases. The system first ranks diseases by their current confidence scores and identifies the symptoms associated with the top diseases that have not yet been queried. If no symptoms are left to ask, the system expands the search to include symptoms from the full knowledge base. The most informative symptom is then selected and justified, ensuring that each question either confirms or rules out high-likelihood diseases, bringing the system closer to a confident diagnosis.

\subsection{Assessment Engine}

\subsubsection{Symptom Matching Phase}

The first step in the assessment engine is to match user-reported symptoms with the knowledge base. Exact or synonym matches contribute their full weight to the condition scores, while semantic matches—determined by cosine similarity between embeddings—contribute 60\%. The similarity is computed as:
\begin{equation}
    \text{sim}(u, s) = \frac{E(u) \cdot E(s)}{\|E(u)\| \cdot \|E(s)\|}
\end{equation}
where \( E(\cdot) \) represents the embedding function for a symptom or user-reported expression.

\subsubsection{Per-Condition Confidence Estimation}

The confidence of each condition hypothesis \( C_d \) is computed through a nonlinear combination of base score, number of matched symptoms, and penalties for denied symptoms:

\begin{equation}
    \text{base\_confidence} = 1 - e^{-0.15 S_d}
\end{equation}

\begin{equation}
    \text{symptom\_factor} = \min\!\left(\frac{\ln(1 + n_d)}{4.0},\; 0.5\right)
\end{equation}

\begin{equation}
    \text{negative\_factor} = \min(0.15 N_d,\; 0.6)
\end{equation}

\begin{equation}
    \text{raw\_conf} = \text{base\_confidence} + \text{symptom\_factor} - \text{negative\_factor}
\end{equation}

\begin{equation}
    C_d = \max\!\big(0,\; \min(0.9 \times \text{raw\_conf},\; 0.95)\big)
\end{equation}

where \( S_d \) is the disease’s accumulated base score, \( n_d \) is the number of matched symptoms, and \( N_d \) is the cumulative penalty from denied symptoms. This formulation caps confidence below 1.0 and enforces smooth growth with diminishing returns.

\subsubsection{Overall Assessment Confidence}

The final assessment confidence \( C_{\text{overall}} \) is computed using the top condition’s score and its margin over the second-highest score:

\begin{equation}
    \text{margin} =
    \begin{cases}
        \dfrac{S_{\text{top}} - S_2}{S_{\text{top}}}, & S_2 > 0 \\[6pt]
        1.0, & \text{otherwise}
    \end{cases}
\end{equation}

\begin{equation}
    \text{raw} = (n_{\text{top}} \times 0.10) + (\text{margin} \times 0.25) + (S_{\text{top}} \times 0.12) - \min(0.05 N_{\text{top}}, 0.5)
\end{equation}

\begin{equation}
    C_{\text{overall}} = \tanh(\text{raw}) \times 0.9
\end{equation}

where \( n_{\text{top}} \) and \( N_{\text{top}} \) denote the number of matched symptoms and denied symptoms penalty for the top-ranked condition respectively.

Logarithmic and exponential transformations were also purposely used to avoid overconfidence, and to come up with more calibrated assessment probabilities. The logarithmic form of expression, which is of the nature of \( \ln(1 + n_d) \), makes sure to maintain diminishing returns; this is that with the increasing number of symptoms matched, the confidence increases less and less, avoiding inflation of confidence in case the number of overlapping or non-specific symptoms matched is large. The base score is stabilized by the exponential term, \( 1 - e^{-0.15S_d} \), which causes the function to saturate with increasing evidence. Lastly, the hyperbolic tangent is used (as the hyperbolic \(\tanh\)) to make the total confidence bounded, so that it can scale smoothly and confidence extreme changes are avoided. A combination of these nonlinearities reduces bias and overconfidence, providing conservative but high-quality assessment confidence estimates.

\subsection{LLM Integration}

\subsubsection{Prompt Templates}

The large language model (LLM) uses specialized prompts to ensure reliable performance across the medical domain. These prompts include:
\begin{enumerate}[i)]
    \item \textbf{Symptom Extraction:} Extracts relevant symptoms from user text using retrieval context.
    \item \textbf{Follow-up Symptom Extraction:}  Identifies new symptoms in user responses to targeted questions.
    \item \textbf{Normalization:}  Standardizes raw symptom phrases into medical terms.
    \item \textbf{Dynamic Question Generation:}  Selects the next most informative symptom to ask about.
    \item \textbf{Test Question Generation:}  Formulates user-friendly questions about assessment tests.
    \item \textbf{Test Result Interpretation:}  Interprets test results against stored thresholds.
    \item \textbf{Risk Factor Evaluation:}  Asks about unaddressed risk factors for specific diseases.
    \item \textbf{Final Assessment Synthesis:}  Combines symptoms, hypotheses, test results, and exclusions into a logical conclusion.
\end{enumerate}

\subsubsection{Chain Orchestration}

It starts with an open-ended elicitation process and then goes over to targeted iterative questioning. The questioning is to be continued until the condition of confirming at least 8 symptoms are fulfilled. Though this cutoff might seem high, the procedure is still feasible since the system has a built-in database of medically relevant terms and synonymous phrases of each symptom and can thus perform a more efficient match and recognition even when users might be using different and informal language.

In case these requirements have been met or the limit of 20 questions have been achieved, the system proceeds into the test evaluation stage. The last  step is a synthesis prompt that integrates the symptoms, test results, and risk factors into a unified assessment report to balance the extent of assessment completeness and assessment accuracy.

\subsection{Introducing Explainability}
This subsection describes the explainability mechanisms integrated into the system, highlighting how assessment reasoning is transparently communicated to users throughout the interaction.
\begin{enumerate}[i)]

\item \textbf{Justification for Symptom Question Selection:}
Figure \ref{symptoms-justification} demonstrates inline justification during symptom questioning, where the system explains why chest pain is being asked in the context of a TB hypothesis, linking the question directly to its assessment rationale.

\begin{figure}[pos=htbp]
\centering
\includegraphics[width=0.9\textwidth]{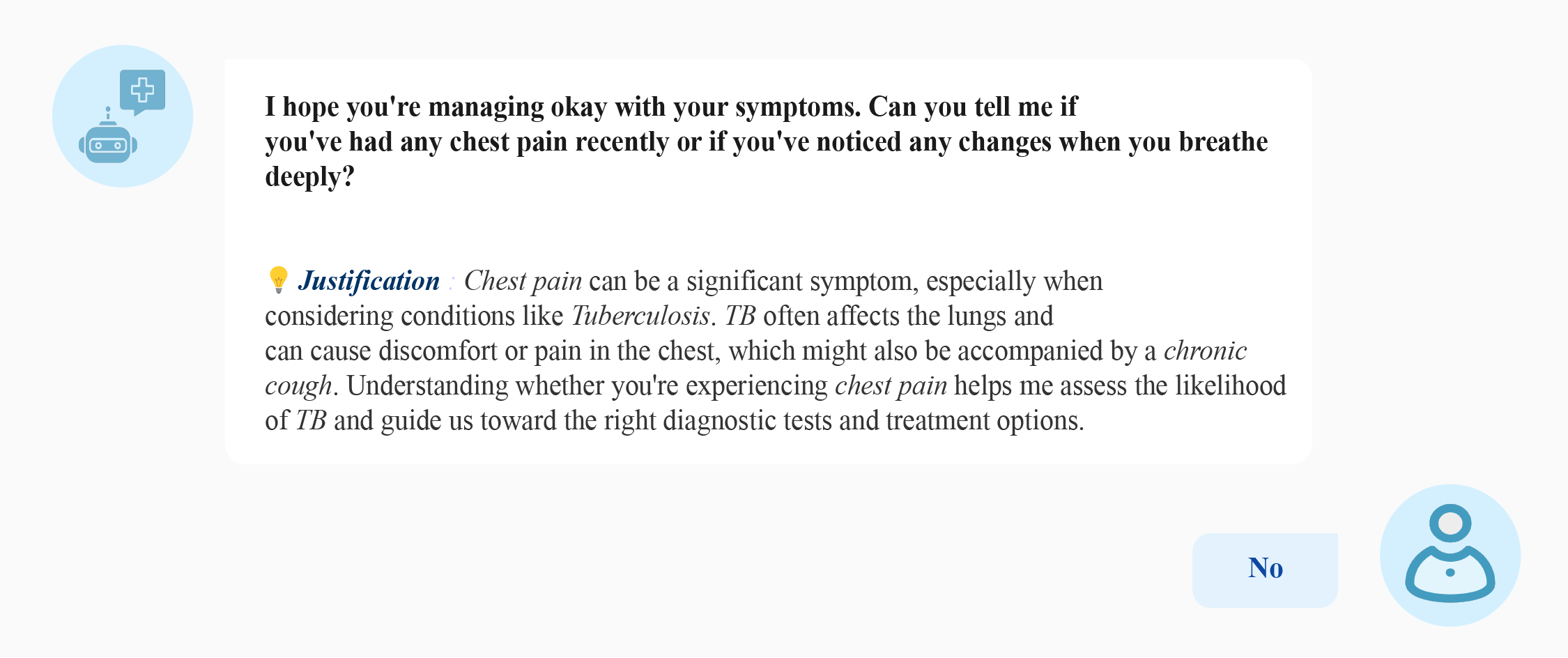}
\caption{User interface showing justification for the selected follow‑up symptom question.}
\label{symptoms-justification}
\end{figure}

\item \textbf{Real-Time Assessment State Updates:}
Figure \ref{disease-status} shows the real-time assessment state after a denied symptom. TB's confidence drops to 19.0\% due to the chest pain penalty (--0.80), while Viral Fever rises to 27.0\%, illustrating how denials actively reshape the condition ranking.

\begin{figure}[pos=htbp]
\centering
\includegraphics[width=0.9\textwidth]{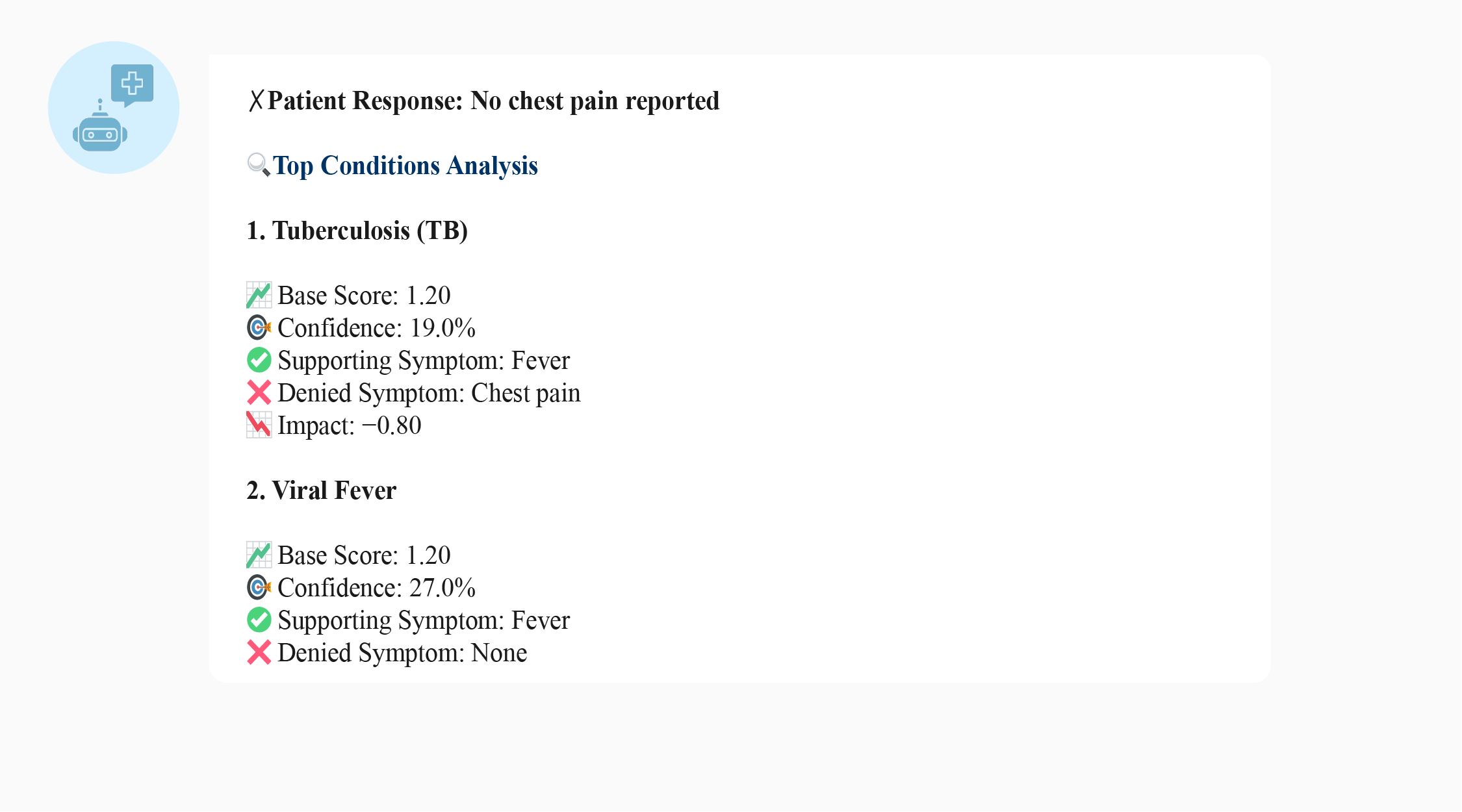}
\caption{User interface showing real-time updates of candidate condition rankings, confidence scores, and accumulated assessment evidence during symptom elicitation.}
\label{disease-status}
\end{figure}

\item \textbf{Test-Based Reasoning and Condition Elimination:}
Figure \ref{test-justification} illustrates test-based condition elimination. A sputum smear result above threshold confirms TB, while a ferritin level of 50 ng/mL exceeds the anemia threshold of 30 ng/mL, resulting in its exclusion.

\begin{figure}[pos=htbp]
\centering
\includegraphics[width=0.9\textwidth]{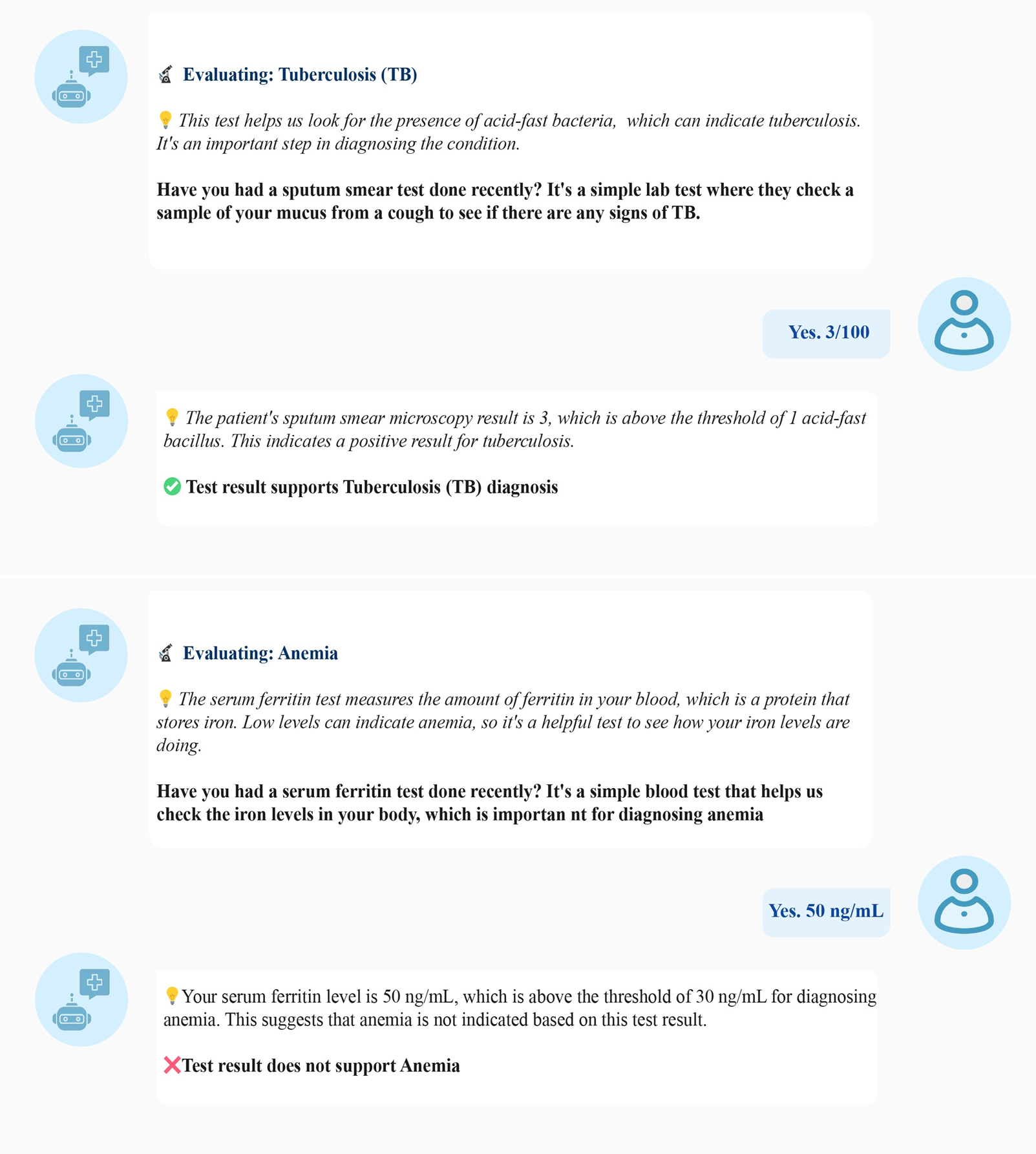}
\caption{User interface illustrating test-based assessment reasoning, including test justification, result interpretation, and elimination of unlikely diseases.}
\label{test-justification}
\end{figure}

\item \textbf{Evidence-Linked Final Assessment and Health Guidance:}
Figure \ref{final-diagnosis-1} shows the final assessment output for a TB case, with supporting evidence drawn from confirmed symptoms, positive sputum smear, and relevant risk factors, alongside explicit reasoning for why competing conditions were deprioritized. Figure \ref{final-diagnosis-2} presents the management recommendations following TB assessment, including treatment regimen, follow-up tests, and specialist referral, completing the system's health-support guidance loop.

\begin{figure}[pos=htbp]
\centering
\includegraphics[width=0.9\textwidth]{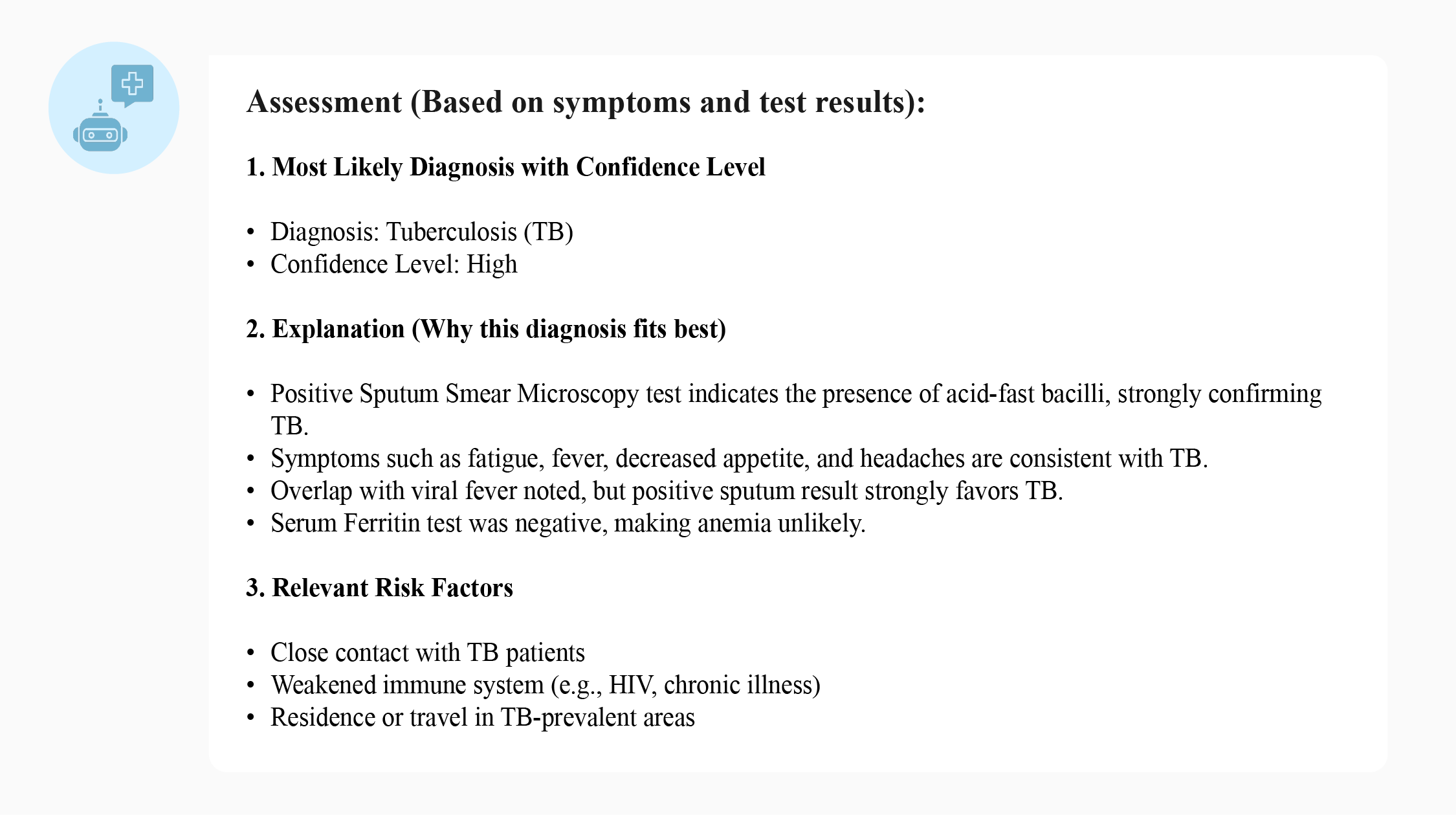}
\caption{Final assessment screen showing the predicted condition together with supporting symptoms, test findings, and risk factors.}
\label{final-diagnosis-1}
\end{figure}

\begin{figure}[pos=htbp]
\centering
\includegraphics[width=0.9\textwidth]{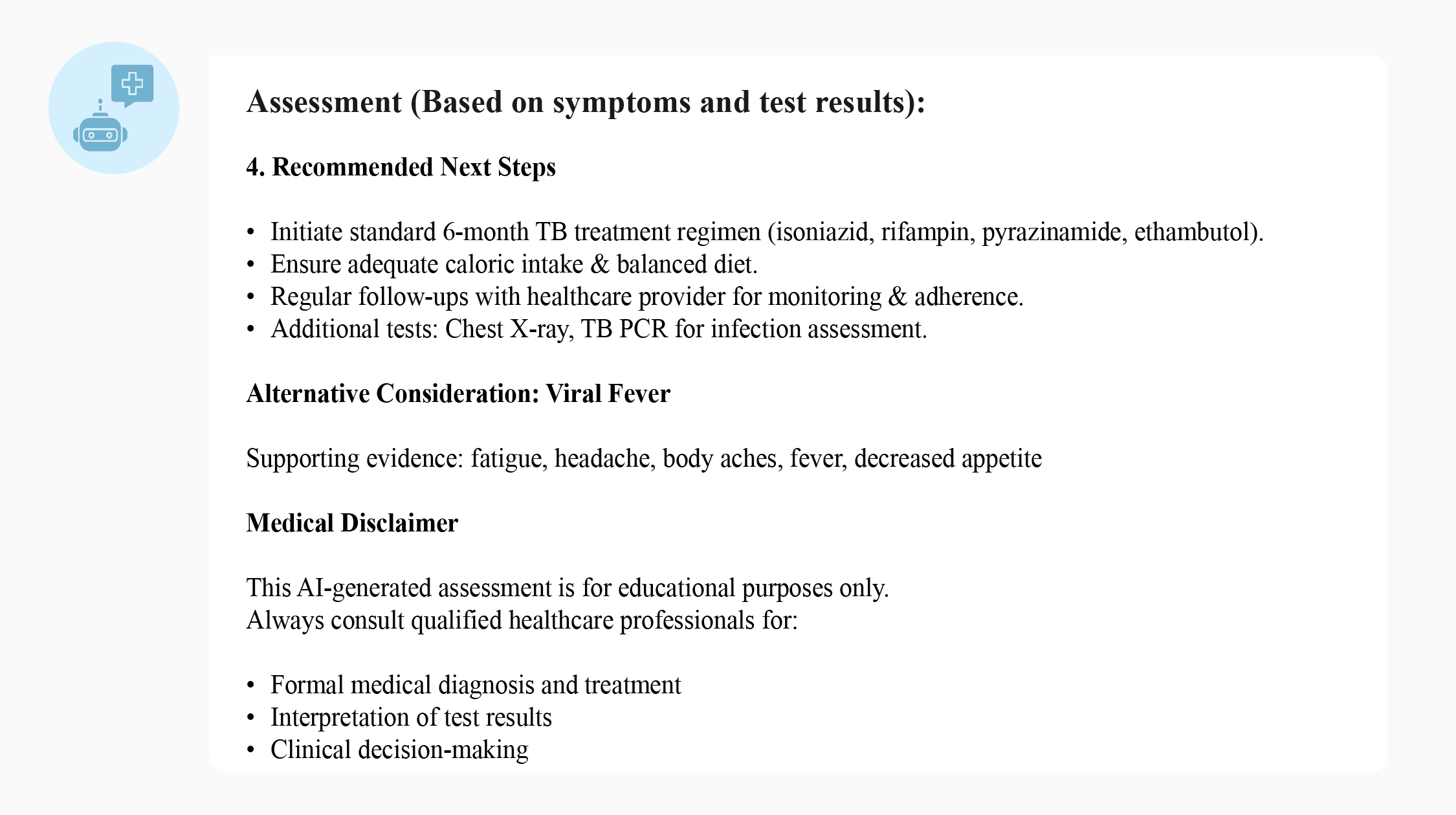}
\caption{Referral and management recommendations generated by the system based on the final assessment.}
\label{final-diagnosis-2}
\end{figure}

\item \textbf{Rationale for the Step-by-Step Approach:}
The explainability design transforms the reasoning process into a transparent and structured interaction that resembles the step-by-step reasoning process commonly followed during a health assessment. All the decisions, including choice of symptoms to entering tests interpretation, are reasoned on each step building trust in the system. This will not only help establish trust in the decisions made by the chat bot, but also enable the validation of the reasoning to be tested and make every exercise consistent during assessment process.

\item \textbf{Condition-Symptom Attribution Map:}

In order to increase the transparency of the assessment reasoning process carried out by our model, a per-condition–symptom attribution visualization is provided in our system for each user-reported symptom, illustrating how it affects the scores of different diseases. Rather than generating a assessment within a black box, the system records which symptoms contribute most heavily to each condition. As the user responds with ``yes'' or ``no'' for symptoms during the conversation, the model continuously updates their respective weights for each of the diseases. Once a symptom is confirmed, the diseases that are frequently associated with that symptom are scored higher.

\begin{figure}[pos=htbp]
    \centering
    \includegraphics[width=\textwidth]{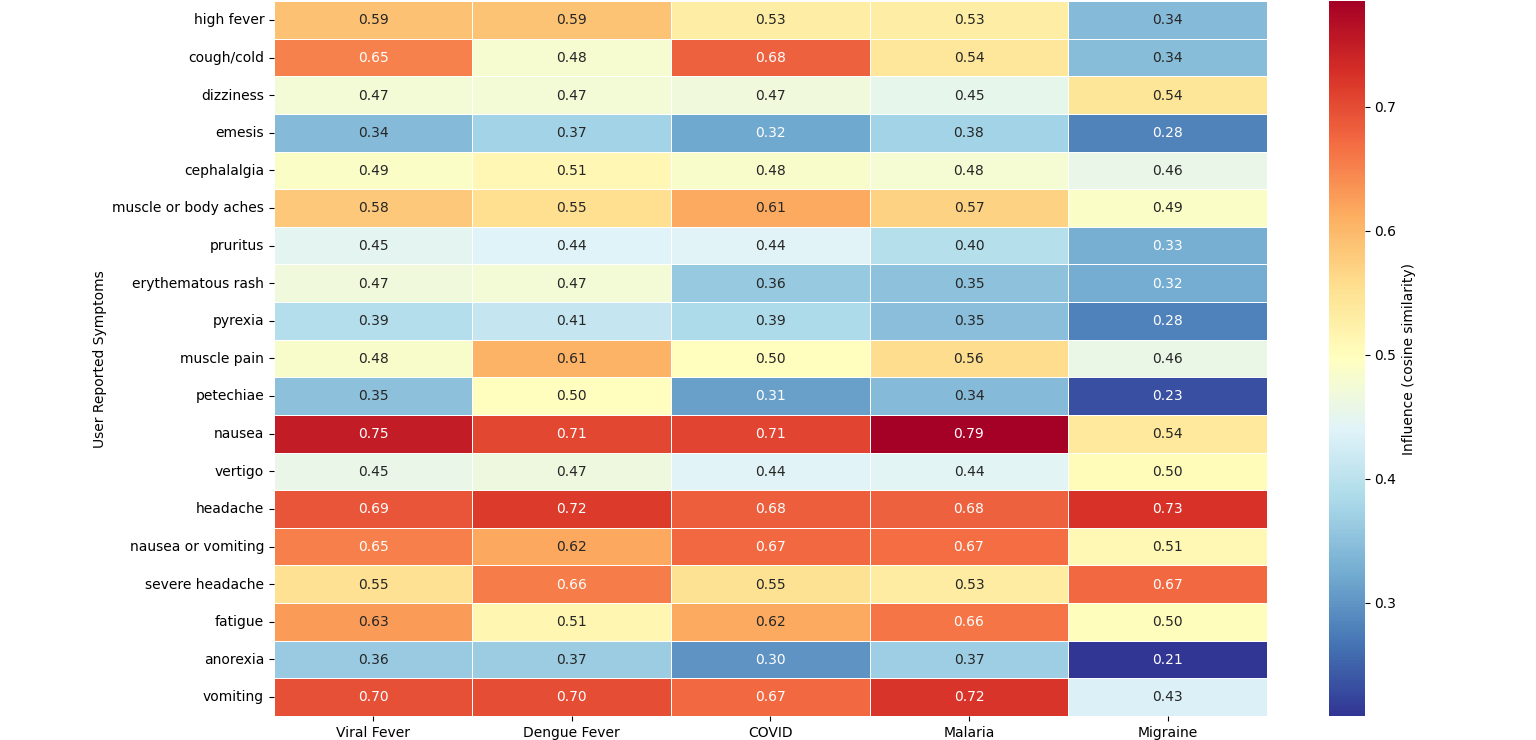}
    \caption{Condition--symptom attribution heatmap showing the contribution of user-reported symptoms (rows) to candidate conditions (columns). Color intensity indicates the relative influence of each symptom on condition scoring.}
    \label{fig:disease_symptom_attribution}
\end{figure}

Figure \ref{fig:disease_symptom_attribution} shows per-symptom assessment influence across candidate diseases. Symptoms such as nausea (0.75 for Viral Fever) and cough (0.68 for Dengue Fever) exhibit high cross-condition influence, while others show more condition-specific patterns, demonstrating how the system weights symptom informativeness during reasoning.

\end{enumerate}
\subsection{Machine Learning Models for Baseline Comparison}
\label{sec:ml_baselines}
We also trained multiple base Machine Learning (ML) models that we could compare the performance of to the proposed LLM-based conversational condition-assessment chatbot. By selecting these models, it was aimed to employ such popular methods of text classification and diagnosing in medicine. The models that were tested has been given:

\begin{enumerate}
    \item Naive Bayes (NB): A simple yet efficient classifier that works well with high-dimensional text data.
    \item Logistic Regression (LR): Used for binary and multi-class classification, assuming linear relationships.
    \item Support Vector Machine (SVM): A margin-based classifier that works well with high-dimensional spaces.
    \item Random Forest (RF): An ensemble method that builds multiple decision trees for more robust predictions.
    \item k-Nearest Neighbors (KNN): A non-parametric method that classifies based on the majority of nearest neighbors.
\end{enumerate}
These models were selected to represent a range of probabilistic, margin-based, ensemble, and instance-based approaches commonly benchmarked in medical text classification tasks. Both TF-IDF and CountVectorizer were used as feature extraction methods to assess how representational choice affects classification performance. In both cases, the vectorizer was configured with English stop word removal and bigram features (ngram\_range = (1,2)) to ensure a fair and consistent comparison across methods. Each model was trained on a single canonical symptom profile per disease, extracted from the curated knowledge base, which is the same knowledge representation used by the LLM system. This setup isolates the contribution of the LLM's conversational reasoning and semantic understanding over feature-based classification given identical knowledge input. All models were evaluated on the same set of user test cases used in the LLM evaluation.

\subsection{Expert Evaluation by Physicians}
\label{sec:clinician}

To evaluate the usefulness, interpretability, and safety of the proposed chatbot in a health-support context, an expert review study was conducted using an observational cross-sectional survey design. The evaluation was carried out using a structured web-based questionnaire after direct interaction with the chatbot. Participants were asked to use the system in a simulated health-support scenario and then complete the survey based on their experience.

\noindent\textbf{Study Setting and Duration.} Physicians were recruited from various hospital settings in Dhaka, Bangladesh. Data collection was conducted between 11 June 2026 and 20 June 2026.

\subsubsection{Eligibility Criteria and Participant Selection}
\begin{figure}[pos=htbp]
    \centering
    \includegraphics[width=0.7\textwidth]{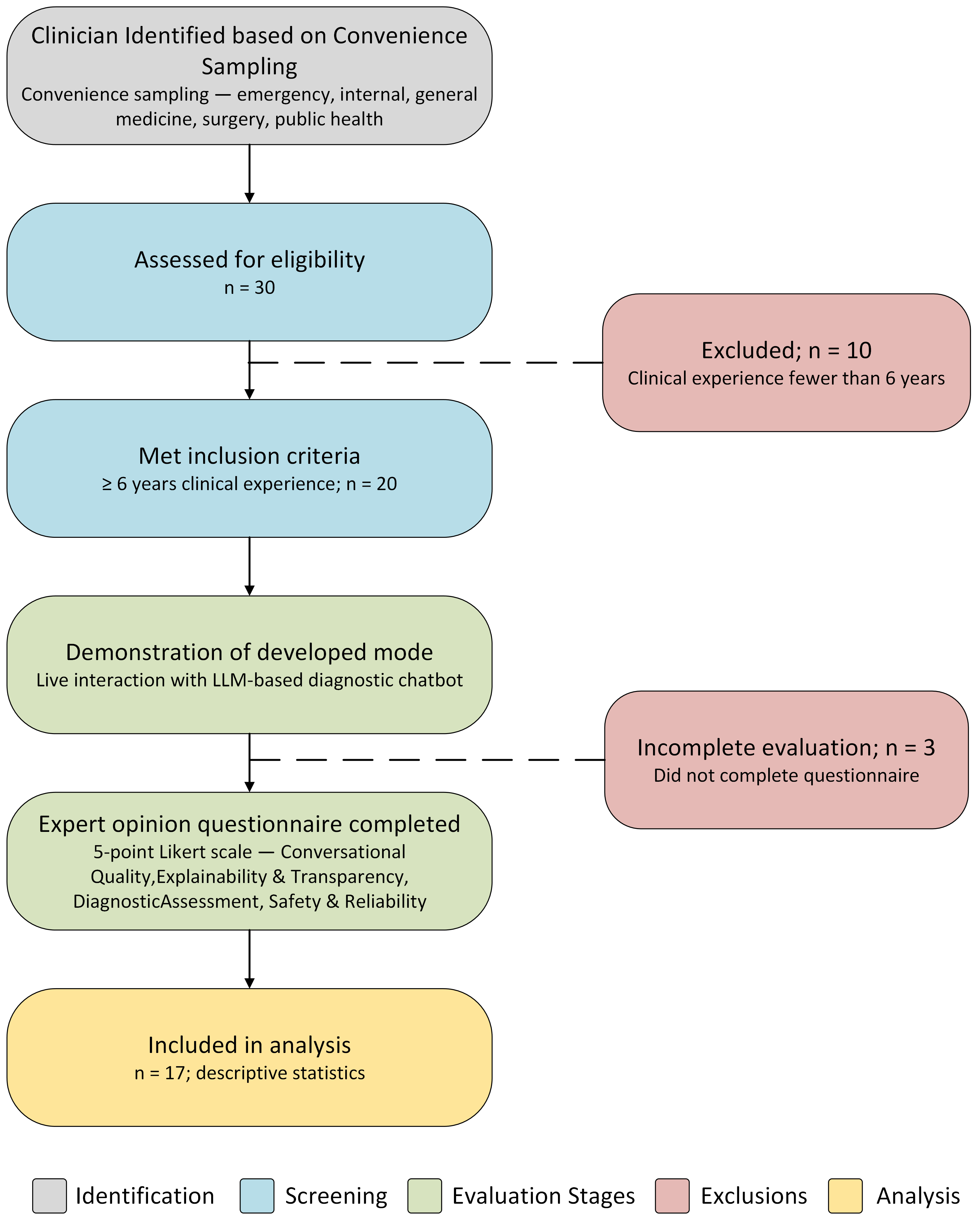}
    \caption{STROBE participant flow diagram for the cross-sectional expert evaluation study (n = 30 assessed; n = 17 analysed)}
    \label{fig:strobe}
\end{figure}

A total of 30 physicians were assessed for eligibility, of whom 10 were excluded for having fewer than 6 years of clinical experience. The remaining 20 physicians met the inclusion criteria and were approached in person, all of whom agreed to participate. Following the system demonstration, 3 participants did not complete the questionnaire and were excluded from analysis, yielding 17 complete responses. This sample size is consistent with expert panel evaluation norms, wherein panels of 10--20 domain experts are considered sufficient for structured clinical usability evaluation studies. Figure \ref{fig:strobe} presents the participant flow diagram for the expert evaluation study.

\subsubsection{Study Participants}
\begin{figure}[pos=htbp]
    \centering
    \includegraphics[width=0.9\textwidth]{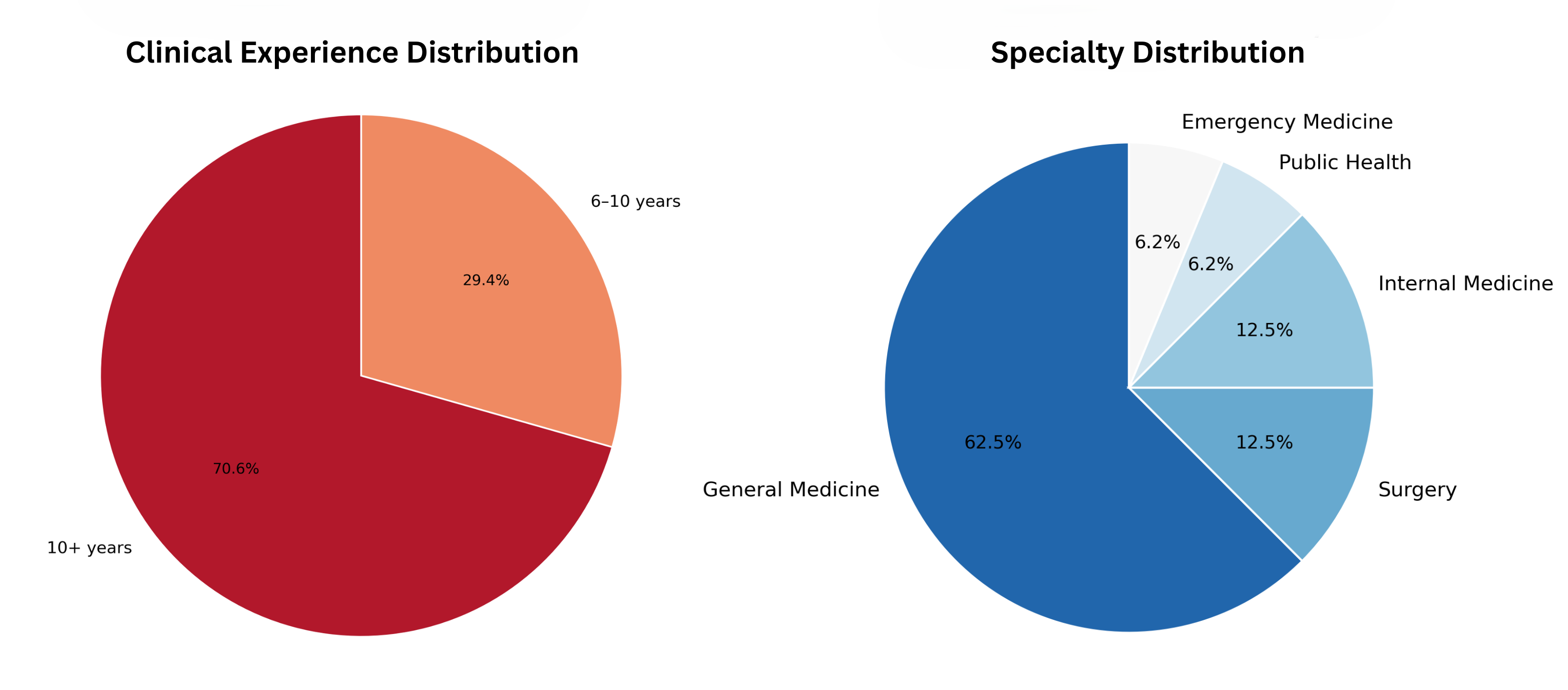}
    \caption{Participant demographics. (a) Specialty distribution of participating physicians (n = 16). (b) Distribution of physicians by years of clinical experience (n = 17).}
    \label{fig:demographics}
\end{figure}

A total of 17 physicians participated in the evaluation, representing diverse specialties and levels of professional experience. Figure \ref{fig:demographics} summarizes the specialty and clinical experience distribution of the participating physicians. Most respondents practiced General Medicine (62.5\%), followed by Internal Medicine and Surgery (12.5\% each), with Public Health and Emergency Medicine each represented by one physician (6.2\%). One respondent did not report a specialty and was therefore excluded from specialty-level analysis only; no Likert response items had missing data across the 17 complete responses. In terms of experience, 70.6\% of physicians had more than 10 years of clinical practice, while the remaining 29.4\% had 6 to 10 years. No physicians with fewer than 6 years of experience took part in the evaluation. No subgroup analyses were conducted for the expert evaluation component due to the small sample size.

\subsubsection{Survey Design}
The questionnaire consisted of 18 items grouped into five domains:
\begin{enumerate}
    \item Conversational Quality
    \item Explainability \& Transparency
    \item Assessment Quality
    \item Safety \& Reliability
    \item Overall Assessment
\end{enumerate}

All items were rated using a 5-point Likert scale (1 = lowest/strongly negative, 5 = highest/strongly positive). The survey assessed multiple aspects of system behaviour, including relevance of follow-up questions for health support, symptom tracking, coherence of multi-turn dialogue, plausibility of the ranked outputs, appropriateness of recommended tests, interpretability of model outputs, and perceived trustworthiness as a health-support tool.

\subsubsection{Data Analysis}
Responses were collected and analysed using descriptive statistics: mean $\pm$ standard deviation (SD), median and interquartile range (IQR), and percentage of positive responses (ratings $\geq 4$). All analyses were performed at both the item and domain levels to summarise physician perceptions of system performance. As convenience sampling was employed, no sampling-weight adjustments were applied to the descriptive analyses. No confounder adjustment was performed, as the study involved descriptive summary of expert opinion ratings rather than epidemiological exposure-outcome analysis. The following statistical measures were computed:

\begin{equation}
Mean \ score (\mu) = \frac{1}{n} \sum_{i=1}^{n} x_i
\end{equation}

where $x_i$ is the Likert score for participant $i$ and $n = 17$.

\begin{equation}
Standard \ deviation\ (SD) = \sqrt{\frac{1}{n-1} \sum_{i=1}^{n} (x_i - \mu)^2}
\end{equation}

\begin{equation}
Interquartile \ Range \  (IQR) = Q_3 - Q_1
\end{equation}

where $Q_1$ and $Q_3$ are the 25th and 75th percentiles of the collected Likert responses, respectively.

\begin{equation}
Percentage \ of \ positive \ responses \ \%(\geq 4) = \frac{\sum_{i=1}^{n} \mathbb{1}(x_i \geq 4)}{n} \times 100
\end{equation}

where $\mathbb{1}(x_i \geq 4)$ equals 1 if participant $i$ gave a rating of 4 or 5, and 0 otherwise.

\noindent\textbf{Potential Sources of Bias.} Several potential sources of bias were identified. Social desirability bias may have influenced physician ratings, as participants were aware they were evaluating a system developed by the research team. Specialty concentration was observed post hoc, with 62.5\% of respondents identifying as general medicine practitioners, reflecting the convenience sampling approach rather than intentional specialty targeting. Additionally, convenience sampling from hospital settings within a single city limits the representativeness of the sample to the broader clinical population.

\section{Results and Discussion}

\subsection{Experimental Objectives and Tasks}

The experiments were designed to evaluate different aspects of the proposed assessment system, including assessment performance, robustness, explainability, and system design choices. The evaluation was divided into several tasks, each addressing a specific research objective.

\begin{enumerate}
    \item \textbf{Assessment Performance Evaluation:}
    To assess the overall assessment capability of the system using Top-1 Accuracy, Top-3 Accuracy, Precision(macro), Recall(macro), and F1-Score(macro). This experiment evaluates whether the system can correctly identify diseases from user-reported symptoms and test information.

    \item \textbf{Cross-Validation and Stability Analysis:}
    To examine the consistency and generalizability of the system across different data partitions using 5-fold cross-validation. This analysis evaluates whether performance remains stable under varying case distributions.

    \item \textbf{Comparison with Traditional Machine Learning Models:}
    To compare the proposed LLM-based system with conventional machine learning approaches, including Naive Bayes, Logistic Regression, Support Vector Machine, Random Forest, and K-Nearest Neighbors. The objective is to determine whether the proposed approach offers advantages over established assessment classification methods.

    \item \textbf{Ablation Study:}
    To investigate the contribution of individual system components and configuration parameters, including symptom thresholds, question limits, similarity thresholds, embedding models, and scoring strategies. This experiment identifies which design choices have the greatest impact on assessment performance.

    \item \textbf{Explainability Assessment:}
    To evaluate the transparency of the assessment process by examining the system's ability to justify symptom selection, present assessment evidence, provide condition ranking updates, and explain final predictions.

    \item \textbf{Physician Expert Review:}
    To assess the health-support relevance and practical usefulness of the system through expert review by physicians. The evaluation focuses on assessment quality, reasoning transparency, appropriateness of follow-up questions, and overall usefulness as a health-support tool.
\end{enumerate}

Together, these experiments provide a comprehensive assessment of the proposed system from predictive, technical, and health-support perspectives.

\subsection{Experimental Setup}

\subsubsection{Evaluation Data}The proposed system was evaluated using a controlled set of synthetic patient cases constructed across 14 diseases in the knowledge base. Each case simulates a user–system interaction consisting of symptom elicitation steps, where each step corresponds to a single symptom query or response.

In total, the evaluation consists of 540 symptom-level interaction steps (conversational turns) distributed across all condition cases. Each synthetic patient case contains approximately 6–7 symptoms expressed in natural, colloquial language, reflecting how users typically describe their condition (e.g., “feeling hot all over” instead of “fever” and "pain in the belly" instead of "stomachache"). The ground truth labels correspond to the predefined condition associated with each synthetic case. These patient cases are stored in a json file. 

\subsubsection{System Configuration}All experiments were conducted on a local machine equipped with an Intel Core i5-12450H processor and an NVIDIA GeForce GTX 1650 GPU, using a retrieval-augmented generation pipeline for inference. The embedding and retrieval components remain fixed across all experiments to ensure consistency, while only the evaluated system parameters vary in ablation settings.

\subsubsection{Evaluation Metrics}

The proposed system was evaluated using both ranking-based and classification-based metrics. Since the primary objective of Phase 1 is to identify and rank the most probable diseases from symptom information, the evaluation focuses on the quality of condition ranking produced during the symptom-based assessment stage. The test-based refinement stage (Phase 2) was not included in the quantitative evaluation because its role is to differentiate among the top-ranked candidate diseases using disease-specific laboratory tests.

\textbf{Top-1 Accuracy} measures the proportion of cases for which the true condition is ranked first by the system:

\begin{equation}
\text{Top-1 Accuracy} =
\frac{\text{Number of cases where the true condition is ranked first}}
{\text{Total number of cases}}
\end{equation}

\textbf{Top-3 Accuracy} measures the proportion of cases for which the true condition appears anywhere within the three highest-ranked predictions:

\begin{equation}
\text{Top-3 Accuracy} =
\frac{\text{Number of cases where the true condition appears in the top three predictions}}
{\text{Total number of cases}}
\end{equation}

In addition to ranking performance, classification performance was evaluated using macro-averaged Precision, Recall, and F1-Score. For these metrics, only the top-ranked condition prediction was considered. Each condition was treated as an individual class, and per-class metrics were calculated before averaging across all diseases.

\textbf{Macro-Precision} is computed as:

\begin{equation}
\mathrm{Macro\text{-}Precision\ (Top-1)} =
\frac{1}{N}
\sum_{i=1}^{N}
\frac{TP_{i}}{TP_{i} + FP_{i}}
\end{equation}

where $N$ is the total number of diseases, $TP_{i}$ denotes the number of true positives for condition $i$, and $FP_{i}$ denotes the number of false positives for condition $i$.

\textbf{Macro-Recall} is computed as:

\begin{equation}
\mathrm{Macro\text{-}Recall\ (Top-1)} =
\frac{1}{N}
\sum_{i=1}^{N}
\frac{TP_{i}}{TP_{i} + FN_{i}}
\end{equation}

where $FN_{i}$ denotes the number of false negatives for condition $i$.

\textbf{Macro-F1 Score} is computed as:

\begin{equation}
\mathrm{Macro\text{-}F1\ (Top-1)} =
\frac{1}{N}
\sum_{i=1}^{N}
\frac{2 \cdot \mathrm{Precision}_{i} \cdot \mathrm{Recall}_{i}}
{\mathrm{Precision}_{i} + \mathrm{Recall}_{i}}
\end{equation}

For all three metrics, only the top-ranked condition prediction was considered when determining $TP_{i}$, $FP_{i}$, and $FN_{i}$.

\subsection{Performance Evaluation of LLM System}

Table \ref{tab:llm_performance} is a summary of the overall system performance of the LLM-based assessment system in all the test folds. The findings show that the model achieves strong assessment accuracy and overall reliability. The system achieves a Top-1 accuracy of 90.3\%, while the relatively low standard deviation (±9.5\%) indicates consistent performance across different evaluation subsets.
 The Top-3 Accuracy achieved 100\%, indicating that the correct assessment was never excluded from the top three conditions predicted by the model.

\begin{figure}[pos=htbp]
\centering
\includegraphics[width=0.9\textwidth]{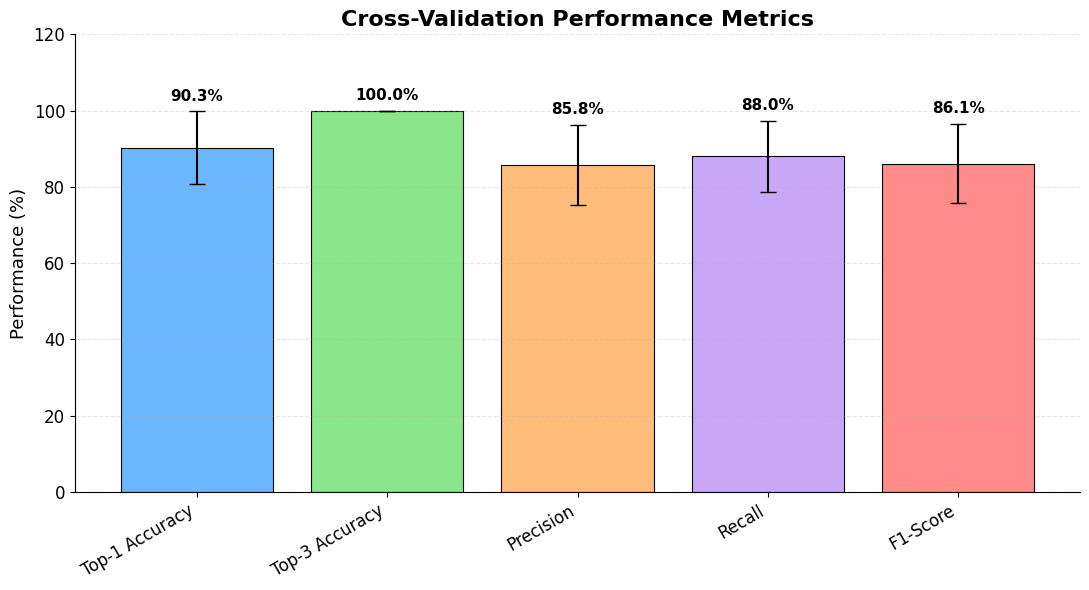}
\caption{Performance metrics of the proposed LLM-based assessment system, including Top-1 Accuracy, Top-3 Accuracy, Macro Precision, Macro Recall, and Macro F1-Score. Error bars indicate standard deviation across the evaluation folds.}
\label{fig:llm_performance}
\end{figure}

The assessment system achieves strong macro precision (85.8\%), macro recall (88.0\%), and macro F1-Score (86.1\%), with moderate variation across folds. The high macro recall indicates that the system successfully identifies most relevant conditions, while the high macro precision demonstrates its ability to reduce false positives. The overall macro F1-Score reflects a balanced performance between macro precision and macro recall. Altogether, Table \ref{tab:llm_performance} shows that the proposed LLM-based system provides accurate assessment predictions across different cases. Figure \ref{fig:llm_performance} visualizes the performance of the proposed LLM-based assessment system across all evaluation metrics, illustrating the balanced nature of its assessment capabilities.

These results can be contextualised against prior work in the field, though direct numerical comparison is constrained by differences in condition scope, dataset origin, and evaluation protocol. Among ML-based approaches operating on text or structured clinical data, \cite{Zhang2026} achieved the most comparable setup, reporting a best Macro F1 of 0.8339 using TF-IDF with logistic regression on multiclass EHR text classification. The proposed system surpasses this with a Macro F1 of 0.861, while additionally supporting free-text input and ranked differential output. Among LLM-based systems, \cite{gupta2025digitaldiagnostics} reported GPT-4 precision of 0.96 and recall of 0.92 on symptom-based diagnosis; however, that system operates via zero-shot prompting without iterative symptom elicitation or confidence-guided ranking, limiting its applicability in realistic assessment workflows. \cite{singhal2022largelanguagemodelsencode} reported 67.6\% accuracy on the MedQA benchmark, though this reflects a medical question-answering task rather than conversational symptom-based diagnosis, making the comparison illustrative rather than direct. Collectively, the proposed system achieves competitive quantitative performance while offering capabilities like ranked predictions, multi-turn symptom elicitation, and explainable confidence scoring, which are absent from the compared baselines.

\begin{table}[htbp]
\centering
\caption{Overall performance of the LLM-based assessment system showing high Top-3 accuracy and balanced macro precision–macro recall performance.}
\label{tab:llm_performance}
\begin{tabular}{l c c c}
\hline
\textbf{Metric} & \textbf{Mean $\pm$ Std. Dev.} & \textbf{Range} & \textbf{Interpretation} \\
\hline
Top-1 Accuracy & 90.3\% $\pm$ 9.5\% & 75.0\% -- 100.0\% & Very High \\
Top-3 Accuracy & 100.0\% $\pm$ 0.0\% & 100.0\% -- 100.0\% & Very High \\
Macro Precision & 85.8\% $\pm$ 10.5\% & 70.8\% -- 100.0\% & Very Good \\
Macro Recall & 88.0\% $\pm$ 9.4\% & 75.0\% -- 100.0\% & Very Good \\
Macro F1-Score & 86.1\% $\pm$ 10.4\% & 72.2\% -- 100.0\% & Very Good \\
\hline
\end{tabular}
\end{table}

\subsection{Fold-wise and Stability Analysis}

Table \ref{tab:fold_performance} presents the fold-wise results of the 5-fold cross-validation experiment, providing further insight into the stability and generalizability of the assessment system. The model demonstrates strong performance across all folds, although some variation is observed. Top-1 Accuracy ranges from 75.0\% to 100.0\%, indicating that while the system performs exceptionally well in most folds, certain patient case distributions are more challenging to diagnose.

Despite these variations in exact-match accuracy, Top-3 Accuracy remains 100\% across all folds, indicating that the correct assessment is consistently included among the top three predictions regardless of the test subset. Macro Precision, Macro Recall, and Macro F1-Score also remain relatively stable across folds, demonstrating consistent performance in identifying and ranking the correct assessments.

The fold-wise analysis further confirms the robustness of the proposed system under varying patient case distributions. The stability of the Top-3 Accuracy and the relatively consistent macro precision and macro recall scores suggest that the model is not overly dependent on a particular subset of the data, supporting its potential applicability in real-world health-support settings.

\begin{table}[htbp]
\centering
\caption{Fold-wise Performance Metrics of the LLM-Based Assessment System (5-Fold Cross-Validation)}
\label{tab:fold_performance}
\renewcommand{\arraystretch}{1.2}
\begin{tabular}{c c c c c c c}
\hline
\textbf{Fold} & \textbf{Cases} & \textbf{Top-1 Accuracy (\%)} & \textbf{Top-3 Accuracy (\%)} & \textbf{Precision} & \textbf{Recall} & \textbf{F1-Score} \\
\hline
1 & 17 & 94.1 & 100.0 & 0.875 & 0.917 & 0.889 \\
2 & 17 & 100.0 & 100.0 & 1.000 & 1.000 & 1.000 \\
3 & 17 & 88.2 & 100.0 & 0.708 & 0.750 & 0.722 \\
4 & 17 & 94.1 & 100.0 & 0.875 & 0.900 & 0.886 \\
5 & 16 & 75.0 & 100.0 & 0.833 & 0.833 & 0.807 \\
\hline
\end{tabular}
\end{table}

\subsection{Condition-wise Performance Analysis}

There was considerable variation in system performance across conditions, providing important insights into the assessment behavior of the proposed approach.

Table \ref{tab:disease_performance} presents disease-wise Top-1 and Top-3 assessment accuracies. Eleven of fourteen conditions achieved 100\% Top-1 accuracy, indicating that the system consistently identified the correct assessment as the most likely outcome for these categories. Their relatively distinctive symptom profiles allow for clear separation in the embedding space, enabling the similarity-based scoring mechanism to discriminate them with high reliability.

Malaria, Hepatitis B, and Viral Fever did not achieve perfect Top-1 accuracy, though all three maintained 100\% Top-3 accuracy, confirming they were always recognized as relevant candidates. Their reduced Top-1 accuracy reflects a well-documented challenge in symptom-based assessment: broad, non-specific symptoms such as fever, fatigue, and headache are shared across many infectious and systemic conditions, making these conditions less distinct in the embedding space. Consequently, competing diseases with more diagnostically specific symptom patterns occasionally receive higher similarity scores, causing the correct condition to be displaced from the top rank. Importantly, this reflects an inherent property of the assessment problem rather than a failure of the retrieval mechanism. Future work incorporating symptom severity weighting and temporal progression features may improve Top-1 disambiguation for these overlapping categories.

\begin{table}[htbp]
\centering
\scriptsize
\caption{Condition Performance Overview Sorted by Accuracy}
\label{tab:disease_performance}
\begin{tabular}{lccc}
\toprule
\textbf{Condition} & \textbf{Top-1 Accuracy (\%)} & \textbf{Top-3 Accuracy (\%)} & \textbf{Unique Symptoms} \\
\midrule

Acute Diarrheal Illness & 100 & 100 & loose stools, fever, loose motions .... \\
Dengue Fever & 100 & 100 & pain behind my eyes, joint pain, retro-orbital pain ....\\
COVID-19 & 100 & 100 & loss of taste/smell, cough, shortness of breath ....\\
Irritable Bowel Syndrome (IBS) & 100 & 100 & distention, diarrhea, gas ....\\
Migraine & 100 & 100 & head pain, sensitivity to light, sensitivity to noise ....\\
Anxiety Disorders & 100 & 100 & irritability, can't sit still, fatigue ....\\
Tuberculosis (TB) & 100 & 100 & coughing up blood, dry cough, cough ....\\
Anemia & 100 & 100 & pale skin, cold hands and feet, difficulty breathing ....\\
Depression & 100 & 100 & thoughts of self-harm, feeling worthless, fatigue ....\\
Asthma & 100 & 100 & chest tightness, coughing a lot, tight feeling in chest ....\\
Gastroesophageal Reflux Disease (GERD) & 100 & 100 & chest pain, burning in chest, belching ....\\

Malaria & 83.33 & 100 & skin turning yellow, fever, chills ....\\
HepatitisB & 50 & 100 & skin turning yellow, exhaustion, fatigue ....\\
Viral Fever & 33.33 & 100 & head pain, aching head, fever ....\\

\bottomrule
\end{tabular}
\end{table}

\subsection{Comparative Analysis with Traditional ML Models}

\begin{figure}[pos=htbp]
\centering
\includegraphics[width=0.9\textwidth, keepaspectratio]{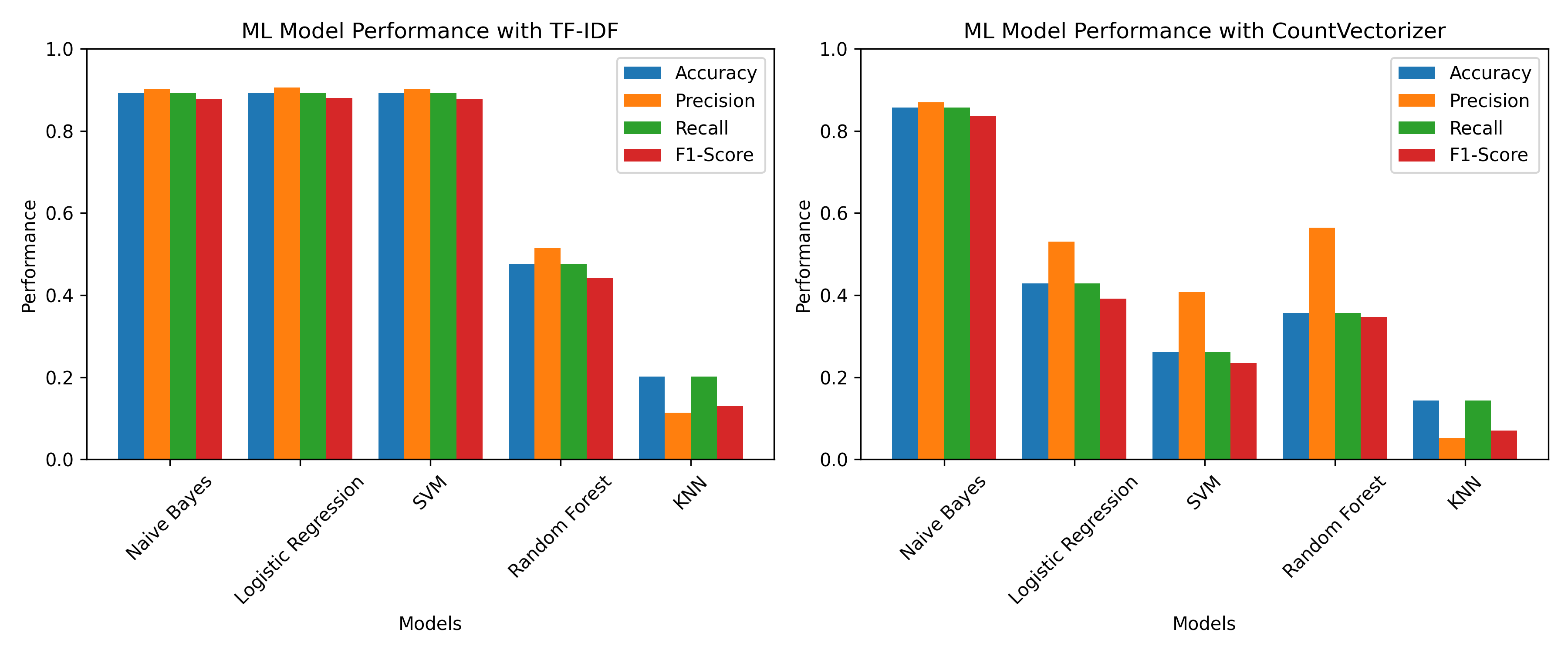}
\caption{Performance comparison of traditional machine learning models using TF-IDF and CountVectorizer feature extraction. Metrics include Accuracy, Macro Precision, Macro Recall, and Macro F1-Score.}
\label{fig:tfidf_count}
\end{figure}

To contextualize the performance of the proposed LLM-based system, a comparative evaluation was conducted against the conventional machine learning models described in Section \ref{sec:ml_baselines}. Five widely used algorithms, namely Naive Bayes, Logistic Regression, SVM, Random Forest, and KNN, were evaluated using TF-IDF and CountVectorizer feature extraction.
TF-IDF consistently outperformed CountVectorizer across all models. Using TF-IDF, Naive Bayes, Logistic Regression, and SVM all achieved comparable performance, with a Macro F1-Score of 0.88 and accuracy of 0.89. Random Forest (F1: 0.44) and KNN (F1: 0.13) performed substantially worse, indicating that these methods are less suited to the single-prototype training setup. With CountVectorizer, Naive Bayes remained the strongest model (F1: 0.84, Accuracy: 0.86), though performance dropped relative to its TF-IDF counterpart. Logistic Regression showed a more pronounced decline (F1: 0.39), and SVM, Random Forest, and KNN all performed poorly under CountVectorizer, demonstrating greater sensitivity to the choice of feature extraction method (Figure~\ref{fig:tfidf_count}).

\begin{figure}[pos=htbp]
\centering
\includegraphics[width=0.9\textwidth, keepaspectratio]{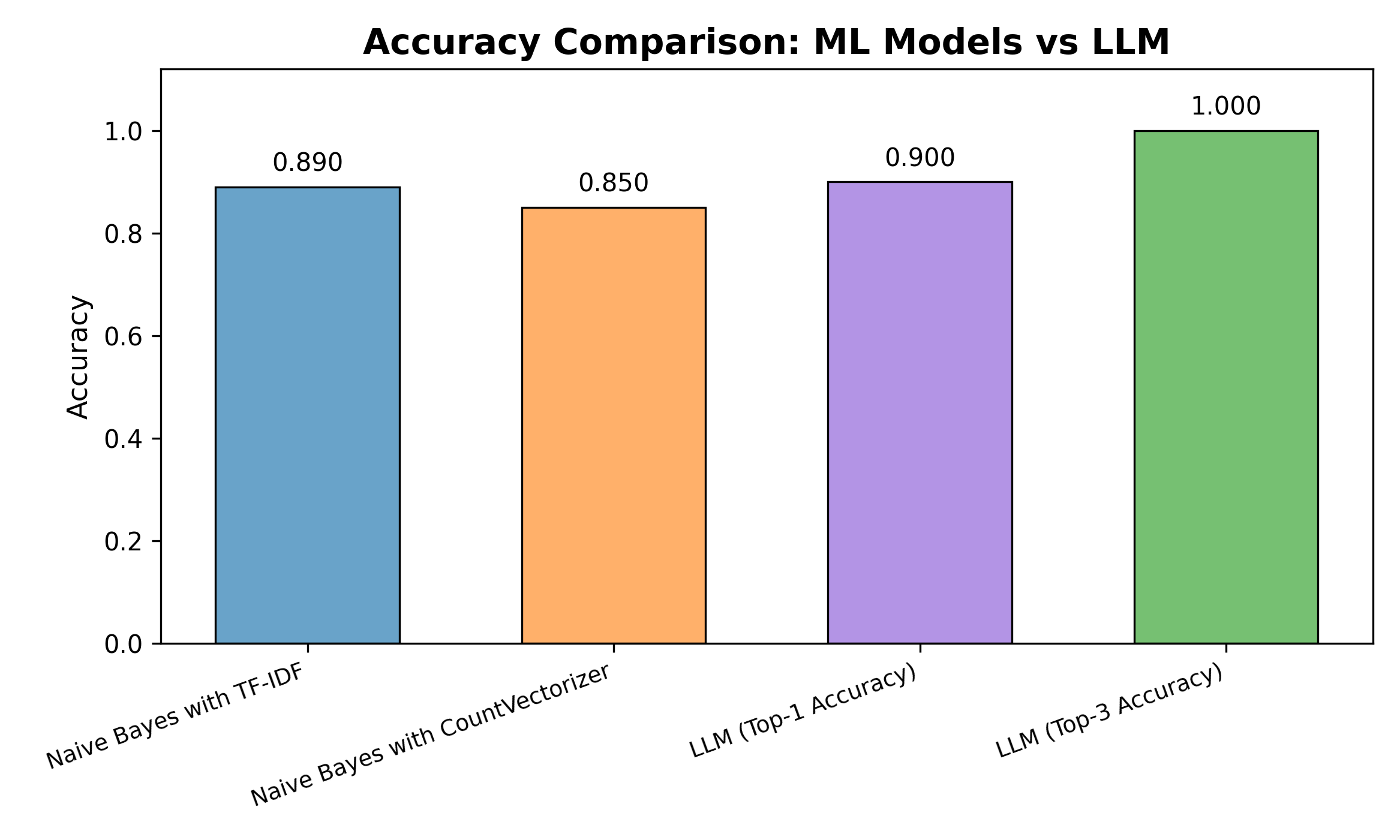}
\caption{Accuracy comparison between the proposed LLM-based assessment system and the best-performing traditional baseline (Naive Bayes) using TF-IDF and CountVectorizer feature extraction.}
\label{fig:ml_count_vs_llm}
\end{figure}

The accuracy metrics in Figure~\ref{fig:ml_count_vs_llm} illustrate the overall performance differences between the traditional models and the proposed LLM-based system. The LLM achieves an accuracy of 0.90, comparable to the best TF-IDF models (0.89) but with a key additional capability: ranked predictions yielding a perfect Top-3 accuracy of 1.00, which traditional single-class models cannot produce by design. The Top-3 metric is particularly relevant in healthcare-related settings, where presenting multiple high-confidence assessment candidates can support decision-making.
In terms of macro precision, the best TF-IDF models (0.90) are marginally higher than the LLM (0.86), indicating slightly more specific positive predictions. Macro recall values are comparable across the LLM (0.88) and the best TF-IDF models (0.89). The macro F1-score shows that TF-IDF Naive Bayes, Logistic Regression, and SVM all achieve 0.88, closely matched by the LLM at 0.86, indicating comparable balance between precision and recall. CountVectorizer Naive Bayes achieves a lower F1 of 0.84, with further degradation across other models.

It is worth noting that the marginal difference in Macro F1 between the proposed system and the best TF-IDF models (0.86 vs. 0.88, $\Delta $=0.02) should be interpreted in context. Traditional models produce a single predicted class and offer no mechanism for uncertainty quantification or conversational follow-up. In contrast, the proposed system achieves perfect Top-3 accuracy of 100\% across all folds, a metric that traditional pipelines cannot produce by design. Furthermore, the traditional baselines operate on clean, pre-segmented symptom inputs, whereas the proposed system processes free-text descriptions with linguistic variation, synonym mapping, and iterative clarification — a considerably harder input condition. Performance parity under these conditions therefore represents a meaningful result in favour of the LLM-based approach.

\subsection{Ablation Study}

\subsubsection{Initial Ablation With All Parameters Enabled (All-MiniLM-L6-v2)}
The first phase of the ablation study was conducted using the All-MiniLM-L6-v2 embedding model while keeping all system parameters enabled. These include the minimum number of questions, minimum symptom requirement, confidence threshold, maximum number of questions, and similarity threshold. The objective was to evaluate how sensitive the assessment system is to individual constraints and which parameters significantly affect predictive behavior. The complete results are presented in Table \ref{table:init_ablation}. Each row varies a single parameter from the baseline configuration while holding all others fixed, isolating the individual effect of that parameter on assessment performance and conversational efficiency.

\begin{table}[H]
\centering
\caption{Initial ablation results using All-MiniLM-L6-v2 with all parameters enabled. MinQ = minimum questions, MinS = minimum confirmed symptoms, Conf = confidence threshold (\%), MaxQ = maximum questions, Sim = similarity threshold, Top1/Top3 = Top-1/Top-3 accuracy, Prec/Rec/F1 = macro precision/recall/F1-score, AvgQ = average questions asked.}
\label{table:init_ablation}
\small
\begin{tabular}{lcccccccccccc}
\hline
Config & MinQ & MinS & Conf & MaxQ & Sim & Top1 & Top3 & Prec & Rec & F1 & AvgQ  \\
\hline
Baseline   & 10 & 8 & 70.00 & 20 & 0.55 & 0.90 & 1.00 & 0.92 & 0.90 & 0.90 & 19.69  \\
MinQ=5     & 5  & 8 & 70.00 & 20 & 0.55 & 0.90 & 1.00 & 0.92 & 0.90 & 0.90 & 19.69  \\
MinQ=15    & 15 & 8 & 70.00 & 20 & 0.55 & 0.90 & 1.00 & 0.92 & 0.90 & 0.90 & 19.69  \\
MaxQ=10    & 10 & 8 & 70.00 & 10 & 0.55 & 0.89 & 1.00 & 0.90 & 0.89 & 0.89 & 10.00 \\
MaxQ=15    & 10 & 8 & 70.00 & 15 & 0.55 & 0.88 & 1.00 & 0.91 & 0.88 & 0.88 & 15.00 \\
MinS=2     & 10 & 2 & 70.00 & 20 & 0.55 & 0.76 & 0.80 & 0.85 & 0.76 & 0.76 & 19.71 \\
MinS=4     & 10 & 4 & 70.00 & 20 & 0.55 & 0.82 & 0.93 & 0.88 & 0.82 & 0.83 & 19.81 \\
MinS=6     & 10 & 6 & 70.00 & 20 & 0.55 & 0.88 & 0.94 & 0.91 & 0.88 & 0.88 & 19.76  \\
Conf=60    & 10 & 8 & 60.00 & 20 & 0.55 & 0.90 & 1.00 & 0.92 & 0.90 & 0.90 & 19.69 \\
Conf=80    & 10 & 8 & 80.00 & 20 & 0.55 & 0.90 & 1.00 & 0.92 & 0.90 & 0.90 & 19.69  \\
Sim=0.45   & 10 & 8 & 70.00 & 20 & 0.45 & 0.88 & 1.00 & 0.90 & 0.88 & 0.87 & 19.74  \\
Sim=0.75   & 10 & 8 & 70.00 & 20 & 0.75 & 0.82 & 0.99 & 0.77 & 0.82 & 0.79 & 19.73  \\
Sim=0.85   & 10 & 8 & 70.00 & 20 & 0.85 & 0.92 & 0.99 & 0.88 & 0.92 & 0.89 & 19.73  \\
\hline
\end{tabular}
\end{table}

The baseline configuration shows excellent performance, with a Top-1 accuracy of 0.90, perfect Top-3 accuracy, and consistent macro precision, macro recall, and macro F1-score. In its baseline behavior, the model typically asks close to twenty questions irrespective of explicit limits, indicating that it attempts to gather a sufficiently rich symptom profile before making a prediction. This baseline provides a clear reference for evaluating the effect of parameter variations.

\textbf{Effect of Maximum Question Limit:}
Changes in the maximum number of questions lead to noticeable behavioral differences. When the system is limited to ten questions, Top-1 accuracy decreases to 0.89 due to insufficient symptom information. A limit of fifteen questions shows a similar trend. This indicates that high-confidence predictions require a sufficient question budget. Although limiting questions reduces conversation length, it can compromise assessment quality.

\textbf{Effect of Minimum Symptom Requirement:}
One of the most influential parameters is the minimum number of confirmed symptoms. Reducing the threshold to two symptoms leads to a sharp drop in accuracy to 0.59. Increasing the threshold to four improves performance significantly, while a threshold of six yields an accuracy of 0.88. These findings indicate that the model relies heavily on symptom depth for accurate inference. Notably, performance stabilizes between six and eight confirmed symptoms, suggesting that a MinS range of 6 to 8 provides high assessment reliability without excessive questioning.

\textbf{Effect of Similarity Threshold:}
Modifying the similarity threshold produces non-linear but expected effects. A low threshold (0.45) leads to overly permissive matching, increased false positives, and reduced macro F1-score. A higher threshold (0.75) becomes overly restrictive and may miss relevant symptom matches. It is notable that 0.85 achieves the highest Top-1 accuracy (0.91), though this comes at a slight cost to macro precision (0.88 versus 0.92 at baseline), suggesting that the stricter threshold occasionally narrows the candidate set in ways that slightly reduce precision even as it improves the single best-match prediction. This confirms that stricter matching is not uniformly beneficial across all metrics, and that 0.55 remains a reasonable balance between Top-1 accuracy and overall macro precision.

\textbf{Confidence Threshold and Minimum Questions:}
Two parameters, namely the minimum number of required questions and the confidence threshold, did not show any consistent effect on assessment accuracy metrics. Across all settings, variations in these parameters resulted in identical Top-1 accuracy, macro precision, macro recall, and macro F1-score. This suggests that the natural behavior of the model already exceeds these thresholds, making them ineffective as control parameters.

Based on this evidence, the next stage of the ablation removed these parameters. This decision was based on empirical evidence, as these parameters did not affect assessment accuracy metrics.

\subsubsection{Refined Ablation Without Confidence Threshold or Minimum Questions}

After setting the confidence threshold to 0 and removing the minimum question constraint, the ablation was repeated with a simplified configuration space. The results (Table \ref{table:final_ablation}) confirm the previous findings. Notably, removing the minimum question constraint allows AvgQ to fall below 19 in several configurations (e.g., MinS=2, MinS=4), revealing that the system terminates conversations earlier when the symptom threshold is relaxed, rather than continuing to question unnecessarily. The maximum question limit continues to influence accuracy, particularly when restricted to ten or fifteen questions. The minimum symptom requirement again shows the strongest effect, with accuracy reaching its lowest point at MinS=2 and improving as stricter requirements are imposed. The stabilization of performance at MinS values between 6 and 8 further supports this as the optimal operating range.

\begin{table}[H]
\centering
\caption{Ablation results after removing the confidence threshold and minimum question constraint (Conf fixed at 0, MinQ unconstrained). Column abbreviations follow Table \ref{table:init_ablation}: MinS = minimum confirmed symptoms, MaxQ = maximum questions, Sim = similarity threshold, Top1/Top3 = Top-1/Top-3 accuracy, Prec/Rec/F1 = macro precision/recall/F1-score, AvgQ = average questions asked.}
\label{table:final_ablation}
\small
\begin{tabular}{lcccccccccccc}
\hline
Config & MinQ & MinS & Conf & MaxQ & Sim & Top1 & Top3 & Prec & Rec & F1 & AvgQ  \\
\hline
Baseline   & 10 & 8 & 0.00 & 20 & 0.55 & 0.90 & 1.00 & 0.92 & 0.90 & 0.90 & 19.69  \\
MaxQ=10    & 10 & 8 & 0.00 & 10 & 0.55 & 0.89 & 1.00 & 0.90 & 0.89 & 0.89 & 10.00  \\
MaxQ=15    & 10 & 8 & 0.00 & 15 & 0.55 & 0.88 & 1.00 & 0.91 & 0.88 & 0.88 & 15.00  \\
MinS=2     & 10 & 2 & 0.00 & 20 & 0.55 & 0.60 & 0.80 & 0.66 & 0.60 & 0.58 & 11.75  \\
MinS=4     & 10 & 4 & 0.00 & 20 & 0.55 & 0.82 & 0.90 & 0.87 & 0.82 & 0.83 & 14.63  \\
MinS=6     & 10 & 6 & 0.00 & 20 & 0.55 & 0.88 & 0.94 & 0.91 & 0.88 & 0.88 & 18.25  \\
Sim=0.45   & 10 & 8 & 0.00 & 20 & 0.45 & 0.88 & 1.00 & 0.90 & 0.88 & 0.87 & 19.74  \\
Sim=0.75   & 10 & 8 & 0.00 & 20 & 0.75 & 0.82 & 0.99 & 0.77 & 0.82 & 0.79 & 19.73  \\
Sim=0.85   & 10 & 8 & 0.00 & 20 & 0.85 & 0.92 & 0.99 & 0.88 & 0.92 & 0.89 & 19.73  \\
\hline
\end{tabular}
\end{table}

The similarity threshold behaves consistently with earlier observations: lower thresholds reduce macro precision due to over-permissive matching, while moderate thresholds yield the most balanced overall performance. The relationship between threshold and accuracy is non-linear; both excessively low and excessively high values degrade at least one metric, supporting the selected value of 0.55 as an empirically validated operating point. Since the confidence threshold and minimum question requirement showed no impact in either ablation stage, this configuration is considered the final validated ablation setup for the All-MiniLM-L6-v2 model.

\subsubsection{Embedding Model Comparison: All-MiniLM-L6-v2 vs. Text-Embedding-Ada-002}

All-MiniLM-L6-v2 was adopted as the default embedding model in the initial system design on the basis of its lightweight, locally-deployable nature, as it runs inference directly on the host machine without requiring external API calls for the embedding step. To assess whether this choice came at a meaningful accuracy cost relative to a stronger commercial alternative, the ablation was repeated using OpenAI's text-embedding-ada-002 model under the same configuration space. The results are presented in Table \ref{table:ada_ablation}.
At the validated operating range (MinS 6--8), the two models perform comparably overall. At the baseline configuration (MinS=8), ada-002 achieves a marginally higher Top-1 accuracy (0.92 vs. 0.90) and Top-3 accuracy (0.99 vs. 1.00 in favour of MiniLM), while MiniLM shows slightly higher macro precision (0.92 vs. 0.88) and macro F1-score (0.90 vs. 0.89). At MinS=6, ada-002 is marginally ahead on Top-1 (0.90 vs. 0.88) and Top-3 (0.96 vs. 0.94), while MiniLM again achieves comparable precision (0.91 vs. 0.93) and F1 (0.88 vs. 0.90). Neither model dominates consistently across all metrics, though MiniLM shows a small but consistent advantage in macro precision and F1 at the selected operating point. Combined with the practical benefit of requiring no additional network calls or per-call costs for the embedding step, independent of the network dependency already introduced by the GPT-4o component, All-MiniLM-L6-v2 was retained as the embedding model for the final system.

\begin{table}[H]
\centering
\caption{Ablation results using the text-embedding-ada-002 model under the same configuration space as Table \ref{table:final_ablation}. Column abbreviations are identical to those defined in Table \ref{table:init_ablation}.}
\label{table:ada_ablation}
\small
\begin{tabular}{lcccccccccccc}
\hline
Config & MinQ & MinS & Conf & MaxQ & Sim & Top1 & Top3 & Prec & Rec & F1 & AvgQ  \\
\hline
Baseline   & 10 & 8 & 0.00 & 20 & 0.55 & 0.92 & 0.99 & 0.88 & 0.92 & 0.89 & 19.79  \\
MaxQ=10    & 10 & 8 & 0.00 & 10 & 0.55 & 0.87 & 0.98 & 0.89 & 0.87 & 0.85 & 10.00  \\
MaxQ=15    & 10 & 8 & 0.00 & 15 & 0.55 & 0.88 & 0.99 & 0.84 & 0.88 & 0.86 & 15.00  \\
MinS=2     & 10 & 2 & 0.00 & 20 & 0.55 & 0.63 & 0.74 & 0.72 & 0.63 & 0.62 & 11.39  \\
MinS=4     & 10 & 4 & 0.00 & 20 & 0.55 & 0.77 & 0.87 & 0.82 & 0.77 & 0.78 & 13.57  \\
MinS=6     & 10 & 6 & 0.00 & 20 & 0.55 & 0.90 & 0.96 & 0.93 & 0.90 & 0.90 & 18.23  \\
Sim=0.45   & 10 & 8 & 0.00 & 20 & 0.45 & 0.92 & 0.99 & 0.88 & 0.92 & 0.89 & 19.79  \\
Sim=0.75   & 10 & 8 & 0.00 & 20 & 0.75 & 0.92 & 0.99 & 0.88 & 0.92 & 0.89 & 19.79  \\
Sim=0.85   & 10 & 8 & 0.00 & 20 & 0.85 & 0.89 & 0.98 & 0.91 & 0.89 & 0.88 & 19.75  \\
\hline
\end{tabular}
\end{table}

\subsubsection{Scoring Weight Ablation}

The final ablation study (Table \ref{table:scoring_weight}) examines different weighting strategies for combining global and local similarity scores in the assessment system. Global similarity scoring evaluates the similarity between a user's confirmed symptoms and the overall symptom profile of each disease. In contrast, local similarity scoring focuses on fine-grained matching between individual user-reported symptoms and disease-specific symptom entries. Global scoring captures overall structural similarity, while local scoring emphasizes precise symptom correspondence. The objective of this ablation is to identify an optimal balance between these two approaches. Table \ref{table:scoring_weight} reports Top-1 and Top-3 accuracy alongside macro precision, recall, and F1-score for each weighting configuration, allowing direct comparison of how the global-local balance affects both classification accuracy and ranking quality.

\begin{table}[H]
\centering
\caption{Scoring weight ablation comparing global-only, local-only, and hybrid similarity strategies. G\_w and L\_w denote the weighting coefficients assigned to global and local similarity scores respectively. Top1/Top3 = Top-1/Top-3 accuracy, Prec/Rec/F1 = macro precision/recall/F1-score.}
\label{table:scoring_weight}
\small
\begin{tabular}{lccccccc}
\hline
Config & G\_w & L\_w & Top1 & Top3 & Prec & Rec & F1 \\
\hline
Local\_Only    & 0.00 & 1.00 & 0.89 & 0.98 & 0.91 & 0.89 & 0.89 \\
Global\_Only   & 1.00 & 0.00 & 0.74 & 0.98 & 0.79 & 0.74 & 0.72 \\
Hybrid\_70\_30 & 0.70 & 0.30 & 0.90 & 1.00 & 0.92 & 0.90 & 0.90 \\
Hybrid\_50\_50 & 0.50 & 0.50 & 0.90 & 0.99 & 0.92 & 0.90 & 0.90 \\
\hline
\end{tabular}
\end{table}

The results show clear differences between scoring strategies. A purely global scoring approach performs worst, with a Top-1 accuracy of 0.7381, indicating that global disease-level similarity alone is insufficient for accurate diagnosis. Local-only scoring performs significantly better, achieving a Top-1 accuracy of 0.8929, demonstrating that fine-grained symptom matching is more diagnostically effective. However, local-only scoring lacks broader contextual information needed to differentiate clinically similar diseases.

These limitations are addressed by hybrid scoring, which combines both global and local similarity. Both the 70/30 (global/local) and 50/50 hybrid configurations achieve high and stable performance, with Top-1 accuracy reaching 0.9048, comparable to or better than the baseline. These hybrid approaches combine the specificity of local scoring with the contextual robustness of global scoring. As a result, hybrid weighting provides the most reliable assessment performance and is selected as the final scoring strategy for the system.

A practical consideration for health-support use is the probability of assessment error. At the validated operating configuration (MinS = 8), Top-3 accuracy reaches 100\% across all folds, indicating that sufficient symptom elicitation consistently yields the correct assessment among the top three candidates. Top-1 errors remain possible, particularly for diseases with overlapping symptom profiles such as Malaria, Hepatitis B, and Viral Fever. The more significant risk arises when the maximum question budget is exhausted before the minimum symptom threshold is met, as reflected in the ablation results at Table \ref{table:final_ablation} MinS=2 (Top-1 = 0.60, Top-3 = 0.80). These findings underscore the importance of treating system output as decision support rather than a definitive diagnosis, with human oversight remaining essential when symptom confirmation is incomplete.

Taken together, the ablation results reveal a hierarchy of parameter sensitivity within the proposed system. The minimum symptom requirement exerts the strongest influence on assessment accuracy, confirming that symptom depth, rather than conversational length alone, is the primary determinant of prediction quality. The similarity threshold produces a non-linear effect, where both overly permissive and overly restrictive values degrade performance, supporting the selected intermediate value of 0.55 as an empirically validated operating point. The maximum question limit shows moderate impact only under severe restriction, while the confidence threshold and minimum question count prove redundant given the model's natural elicitation behaviour. The scoring weight ablation further establishes that neither global nor local similarity alone is sufficient; their combination is necessary for robust performance, with both the 70/30 and 50/50 hybrid configurations outperforming either pure strategy. This hierarchy of sensitivity has practical implications: future system adaptations or deployments should prioritise tuning the minimum symptom requirement and similarity threshold, as these parameters govern the most consequential trade-offs between assessment accuracy and conversational efficiency.

\subsection{Physician Expert Review Results}
This subsection presents the results of the physician expert review, in
which 17 physicians assessed the chatbot across the 14 conditions in
its scope. The findings are organized into participant demographics, domain-wise rating results, the overall distribution of responses, and a summary of the main takeaways.

\subsubsection{Domain-wise Evaluation Results}

Table \ref{tab:physician_survey} reports the mean, median, and percentage of favorable ratings (rating of 4 or higher) for each survey item, grouped by evaluation domain.

\begin{table}[H]
\centering
\caption{Physician survey results by domain (n = 17). The definition of each metric has been defined in Section \ref{sec:clinician}}
\label{tab:physician_survey}
\small

\begin{tabular}{llccc}
\hline
Domain & Item & Mean $\pm$ SD & Median (IQR) & \% $\geq$ 4 \\
\hline
Conversational Quality & Relevance of follow-up questions & 4.29 $\pm$ 0.47 & 4 (4--5) & 100.0 \\
Conversational Quality & Confirmed vs. denied symptom distinction & 4.29 $\pm$ 0.47 & 4 (4--5) & 100.0 \\
Conversational Quality & Multi-turn conversation coherence & 4.12 $\pm$ 0.60 & 4 (4--4) & 88.2 \\
Conversational Quality & Progressive, meaningful questioning & 4.12 $\pm$ 0.49 & 4 (4--4) & 94.1 \\
Explainability and Transparency & Usefulness of symptom display & 4.06 $\pm$ 0.43 & 4 (4--4) & 94.1 \\
Explainability and Transparency & Usefulness of top-3 condition and confidence display & 4.12 $\pm$ 0.49 & 4 (4--4) & 94.1 \\
Explainability and Transparency & Logic and justification of explanations & 4.24 $\pm$ 0.44 & 4 (4--4) & 100.0 \\
Assessment Quality & Symptom registration and retention accuracy & 4.12 $\pm$ 0.33 & 4 (4--4) & 100.0 \\
Assessment Quality & Plausibility of top condition predictions & 4.12 $\pm$ 0.49 & 4 (4--4) & 94.1 \\
Assessment Quality & Appropriateness of recommended tests & 4.12 $\pm$ 0.60 & 4 (4--4) & 88.2 \\
Assessment Quality & Incorporation of test results into assessment & 4.12 $\pm$ 0.49 & 4 (4--4) & 94.1 \\
Assessment Quality & Reasonableness of final assessment & 4.18 $\pm$ 0.39 & 4 (4--4) & 100.0 \\
Assessment Quality & Appropriateness of management recommendations & 4.12 $\pm$ 0.33 & 4 (4--4) & 100.0 \\
Safety and Reliability & Avoidance of unsafe or misleading recommendations & 4.35 $\pm$ 0.49 & 4 (4--5) & 100.0 \\
Safety and Reliability & Trustworthiness as a health-support assistant & 4.29 $\pm$ 0.47 & 4 (4--5) & 100.0 \\
Safety and Reliability & Alignment with health-guidance principles & 4.41 $\pm$ 0.51 & 4 (4--5) & 100.0 \\
Overall Assessment & Overall system rating & 4.35 $\pm$ 0.49 & 4 (4--5) & 100.0 \\
Overall Assessment & Differential assessment narrowing across turns & 4.06 $\pm$ 0.43 & 4 (4--4) & 94.1 \\

\hline
\end{tabular}

\end{table}

Overall, physicians reported consistently favorable evaluations across all assessed domains, with item-level means ranging narrowly from 4.06 to 4.41. Safety and reliability related items were rated the most consistently positive of the five domains. Multi-turn conversation coherence and the appropriateness of recommended assessment tests were still rated favorably overall, but they drew comparatively more hesitation from a small number of physicians.

\subsubsection{Response Distribution Analysis}

\begin{figure}[pos=htbp]
    \centering
    \includegraphics[width=0.9\textwidth]{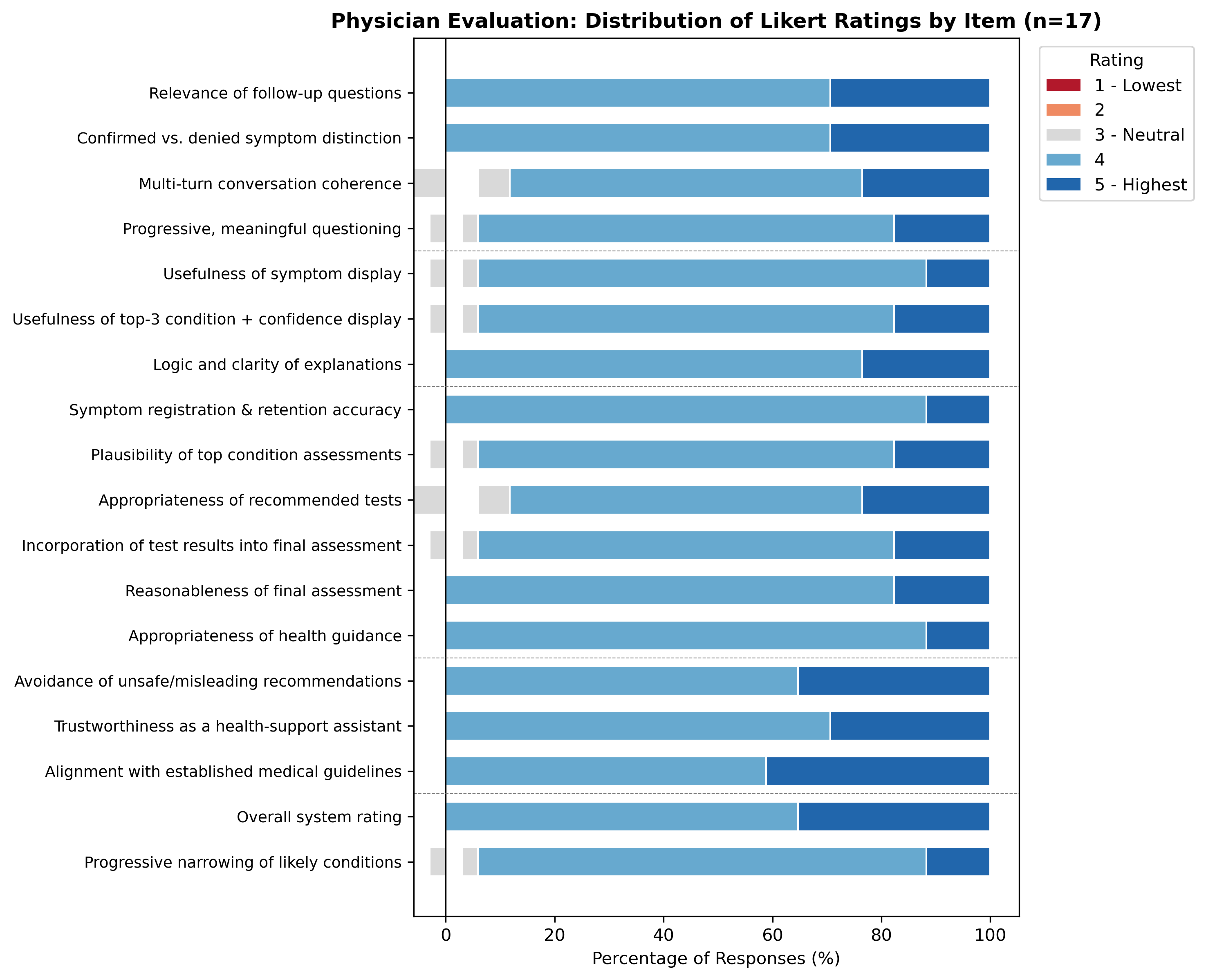}
    \caption{Distribution of physician Likert ratings across all survey items, grouped by domain.}
    \label{fig:likert}
\end{figure}

Figure \ref{fig:likert} illustrates the distribution of physician responses across all 18 survey items.

Ratings were concentrated almost entirely in the categories of 4 and 5. Across all 306 individual ratings (18 items rated by 17 physicians), none fell below a rating of 3, and only 10 ratings (3.3\%) were neutral. This points to strong and consistent agreement regarding the system's usefulness in health-support contexts.
\subsubsection{Key Findings Summary}

The evaluation results show that the system performs well in sustaining coherent, multi-turn conversations and in producing assessment reasoning that physicians found interpretable and plausible. Follow-up question relevance and the distinction between confirmed and denied symptoms were particularly well received, with all physicians rating both items 4 or higher. This suggests that the system's structured symptom-tracking approach was clearly perceived by the participating physicians. Safety related items, including avoidance of unsafe recommendations and alignment with health-guidance principles, were also rated favorably, and no physician rated any item below an acceptable level across the entire evaluation.

When asked specifically about future applicability in healthcare-related
contexts, physician responses were more measured. 35.3\% endorsed the system's potential outright by selecting "Yes," while the remaining 64.7\% selected "Possibly." No physician selected "No." Taken together with the consistently positive domain ratings, this pattern suggests cautious but genuine acceptance of the system among participating physicians as a health-support and pre-screening aid, rather than unconditional support for autonomous assessment.

\subsection{Addressing Challenges of Prior Work}

\subsubsection{Input Flexibility \& Conversational Interaction}

Many ML/DL assessment models \cite{Rahman2025,Tachicart2025,Nazir2023,Zhang2026,Park2021} are designed primarily for structured inputs, and are not intended to process free-text symptom descriptions as naturally expressed by user. Some LLM-based systems \cite{singhal2022largelanguagemodelsencode,gupta2025digitaldiagnostics,mdagents2024} focus on single-turn or static interaction paradigms, which is appropriate for their intended benchmarks but does not extend to iterative, stateful symptom elicitation across multiple turns.
The proposed system is designed specifically for the multi-turn conversational setting, retaining confirmed and denied symptoms across interaction turns and using this accumulated context to guide subsequent questioning. Linguistic normalization and synonym mapping improve robustness to natural variation in user phrasing, supporting consistent symptom recognition across diverse expressions of the same underlying complaint. The dialogue is additionally designed to maintain an empathetic and user-centric tone throughout the interaction.

\subsubsection{Explainability \& Interpretability}
Explainability in conversational assessment systems remains an open challenge. AMIE \cite{tu2024conversationaldiagnosticai} demonstrates strong assessment performance but is not primarily designed to expose how individual symptoms contribute to specific predictions, which is one consideration relevant to clinician oversight and trust.
The proposed system incorporates explainable reasoning at each stage of the assessment interaction. The rationale behind each follow-up question is communicated to the user, clarifying what assessment information is being sought. Upon reaching a assessment conclusion, confirmed symptoms and elicited findings are linked to the predicted conditions, providing a traceable reasoning chain. At each turn, the top three most probable assessments are presented alongside their confidence scores and rankings, allowing users to observe how the assessment distribution evolves as new information is incorporated.
\subsubsection{Reliability \& Hallucination Control}

Prior works such as \cite{clinicalcamel2023, mdagents2024} have reported risks of hallucination in similar systems. Medical LLMs can exhibit bias, misinterpret user inputs, and generate hallucinated or unsupported outputs, which may compromise assessment safety.

To improve output reliability, the system employs grounded generation, embedding-space semantic filtering, and active rejection of unverified outputs, ensuring that generated suggestions remain consistent with the knowledge base. Reliability and uncertainty are managed through a dynamic confidence scoring system, where confirmed symptoms increase the confidence of relevant diseases, and denied symptoms reduce the confidence of diseases that include them. Confidence values are further adjusted using nonlinear scaling, which prevents overconfident predictions and stabilizes ranking behavior. Weighted confidence scoring reduces overemphasis on low-specificity symptoms, such as headache or mild fever.

\subsection{Limitations and Future Work}

While the model demonstrates strong performance metrics and effective hallucination mitigation, it is not yet ready for real-world deployment in healthcare-related contexts. Our approach achieves strong performance on synthetic datasets, but these results should be interpreted with caution. Synthetic data enables controlled and scalable evaluation but does not fully capture the variability and unpredictability of
real-world environments. Consequently, performance gains observed here may not directly translate to practical settings. The generalisability of findings is further limited by the survey's single-city recruitment from Dhaka, Bangladesh, specialty concentration among participants, and the use of convenience sampling. Even though the evaluation was conducted with physician participants, the relatively small sample size (n = 17) limits the generalizability of the findings. Moreover, the data generation process may introduce implicit biases that favor certain modeling assumptions. Future work should validate these findings on real-world data and examine robustness under distributional shifts. Fixed symptom weights limit the model's ability to adjust for symptom severity or contextual relevance in complex, multi-disease, or rare-case scenarios. Although chain-of-thought reasoning provides interpretability, its faithfulness remains unverified and may not accurately reflect the model's internal decision process, which could
influence how users interpret and trust the generated explanations. Additionally, the use of third-party LLMs raises privacy and data security concerns, as sensitive health information may be processed outside secure environments. These limitations have important implications for user safety and the ethical deployment of AI-assisted
health-support systems, highlighting the need for careful validation, appropriate safeguards, and continued human oversight before use in routine healthcare practice. Future work should focus on:

\begin{itemize}
\item Expanding condition coverage to include rare and complex conditions.
\item Improving symptom significance to better capture symptom severity and health-support relevance.
\item Incorporating multimodal large language models capable of processing assessment inputs such as medical images and physiological signals, together with comparisons with more recent medical LLM-based systems, would further improve the robustness and comprehensiveness of the study.
\item Conducting prospective expert evaluations using real-world health data and involving realistic user--professional interactions in healthcare settings to further assess system reliability and applicability in healthcare-related contexts.
\item Strengthening user privacy and data security mechanisms through approaches such as federated learning, secure on-device processing, or privacy-preserving retrieval strategies could further reduce the exposure of sensitive medical data during model interaction.
\end{itemize}

Addressing these areas will enhance the system's reliability, interpretability, and potential for applicability in healthcare-related contexts, moving closer toward safe real-world deployment. While it is not yet ready for routine healthcare practice, its contribution to safe and evidence-based AI-assisted health support remains significant.
\section{Conclusion}

We introduce a real-time, explainable conversational AI system for early-stage health assessment using a large language model. The system combines GPT-4o, retrieval-augmented generation, and chain-of-thought reasoning to facilitate interpretable and interactive health-support dialogue. Evaluation demonstrates strong performance in handling complex conversational assessment scenarios compared to traditional machine learning models. Although the system shows promise, it is not yet ready for real-world deployment in healthcare-related contexts due to inherent limitations of LLM-based approaches. Nevertheless, this work highlights the potential of LLM-based systems to provide safe, reliable, and accessible AI-assisted health support. With further development, such systems could play a significant role in enhancing assessment accuracy and usability.

\section*{Declaration of Generative AI and AI-Assisted Technologies}

During the preparation of this manuscript, the author(s) used \textit{ChatGPT} to generate synthetic patient cases and to assist with phrasing and grammar refinement. Following the use of this tool, the author(s) reviewed and edited all content as necessary and take full responsibility for the accuracy, integrity, and originality of the published work.

\section*{Ethics and Consent}

Ethical approval for this study was obtained from the Biosafety, Biosecurity and Ethical Committee, Faculty of Biological Science, Jahangirnagar University, Bangladesh, prior to data collection (Ref: BBEC.JU/IIT/2026/14; Date: 10 June 2026). The study involved human participants in the form of licensed physicians who completed a structured questionnaire via Google Forms.
The participating physicians anonymously provided responses based on their professional experience. All participants were informed that consent to participate in the study and publish their data would be assumed on completion and submission of the study questionnaire. This observational study was done following the STROBE guidelines.

\section*{Declaration of Interests}
The authors declare that they have no known competing financial interests or personal relationships that could have appeared to influence the work reported in this paper.

\section*{Funding}
This research did not receive any specific grant from funding agencies in the public, commercial, or not-for-profit sectors.

\section*{Author Contributions}
Maliha Tabassum: Conceptualization; Methodology; Software; Validation; Formal analysis; Investigation; Data curation; Writing – original draft; Writing – review \& editing; Visualization.

M. Shamim Kaiser: Supervision; Conceptualization; Methodology; Writing – review \& editing.

\section*{Data and Code Availability}
This study did not use any publicly available datasets. The knowledge base was constructed by curating information from authoritative publicly accessible sources, including guidelines and reports published by the World Health Organization (WHO) and other established medical organizations, all of which are cited in the manuscript. The source code for the proposed system, including the web-based implementation, model execution, and evaluation scripts, is available from the corresponding author upon reasonable request.

\bibliographystyle{plain}
\bibliography{cas-refs}
\end{document}